\documentclass[11pt]{article}

\usepackage{acl}

\usepackage{times}
\usepackage{latexsym}
\usepackage{booktabs}
\usepackage[T1]{fontenc}
\usepackage{courier}
\usepackage{tabularx}
\usepackage[utf8]{inputenc}
\usepackage{multirow}
\usepackage{microtype}
\usepackage{amssymb}

\usepackage{inconsolata}
\usepackage{amsmath}

\usepackage{graphicx}
\usepackage{subcaption}
\usepackage{threeparttable}
\usepackage[most]{tcolorbox}

\newtcolorbox{promptbox}{
    enhanced,
    breakable,
    colback=blue!5!white,
    colframe=blue!75!black,
    fonttitle=\bfseries,
    rounded corners,
    boxrule=1pt,
    arc=4pt,
    left=6pt,
    right=6pt,
    top=6pt,
    bottom=6pt,
}
\title{Safe-SAIL: Towards a Fine-grained Safety Landscape of Large Language Models via Sparse Autoencoder Interpretation Framework
\\
{\color{red} \small Warning: this paper contains data, prompts, and model outputs that are offensive in nature.}}

\author{
    \stepcounter{footnote}
    Jiaqi Weng$^{1}$\footnotemark[1],\;
    Han Zheng$^{2}$\footnotemark[1],\;  
    Hanyu Zhang$^{1}$,\;
    Ej Zhou$^{3}$,\;
    Qinqin He$^{1}$,\;
    Jialing Tao$^{1}$\footnotemark[2],\; \\
    \textbf{Hui Xue$^{1}$},\;
    \textbf{Zhixuan Chu$^{2}$}\footnotemark[2],\;
    \textbf{Xiting Wang$^{4}$}\\
    \normalsize{$^{1}$Alibaba Group ~$^{2}$The State Key Laboratory of Blockchain and Data Security, Zhejiang University}\\
~~\normalsize{$^{3}$Language Technology Lab, University of Cambridge ~$^{4}$Renmin University of China 
} \\ 
\texttt{\{wengjiaqi.wjq, jialing.tjl\}@alibaba-inc.com, \{h.zheng\}@zju.edu.cn}
}

\definecolor{ejcolor}{RGB}{0,120,180}

\begin{document}
\maketitle


\begin{abstract}

\renewcommand{\thefootnote}{\fnsymbol{footnote}}
\footnotetext[1]{Equal contribution}
\footnotetext[2]{Corresponding author}
\renewcommand{\thefootnote}{\arabic{footnote}}
\setcounter{footnote}{0}

Sparse autoencoders (SAEs) enable interpretability research by decomposing entangled model activations into monosemantic features.  However, under what circumstances SAEs derive most fine-grained latent features for safety—a low-frequency concept domain—remains unexplored. 
Two key challenges exist: identifying SAEs with the greatest potential for generating safety domain-specific features, and the prohibitively high cost of detailed feature explanation. In this paper, we propose \textbf{Safe-SAIL}, a unified framework for interpreting SAE features in safety-critical domains to advance mechanistic understanding of large language models. 
Safe-SAIL introduces a pre-explanation evaluation metric to efficiently identify SAEs with strong safety domain-specific interpretability, and reduces interpretation cost by 55\% through a segment-level simulation strategy.
Building on Safe-SAIL, we train a comprehensive suite of SAEs with human-readable explanations and systematic evaluations for 1,758 safety-related features spanning four domains: pornography, politics, violence, and terror. Using this resource, we conduct empirical analyses and provide insights on the effectiveness of Safe-SAIL for risk feature identification and how safety-critical entities and concepts are encoded across model layers.
All models, explanations, and tools are publicly released in our open-source toolkit\footnote{\url{https://github.com/Alibaba-AAIG/Safe-SAIL}} and companion product\footnote{\url{https://modelscope.cn/studios/Alibaba-AAIG/Safe-SAIL/summary}}.
\end{abstract}

\section{Introduction}

\begin{figure}
\centering
\includegraphics[width=\columnwidth]{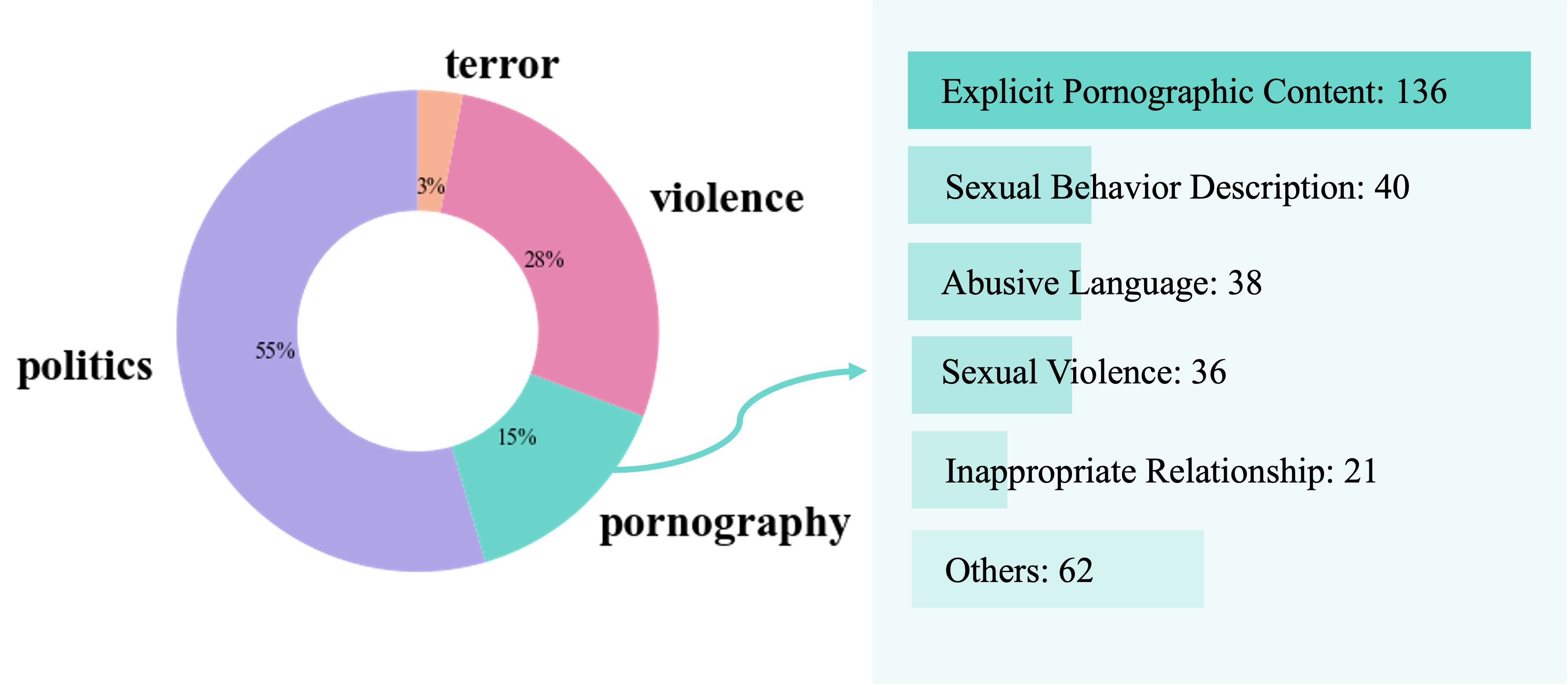}
\caption{Overview of our trained safety‑related SAE feature-base. The feature-base covers four safety domains; here we show the pornography domain as an example. In total, 309 features from SAEs trained on quarter layers (0, 8, 17, 26, 35) are interpreted as related to pornography. The chart on the right lists the top five sub‑categories of these features. 
}
\label{overview}
\end{figure}

Increasing deployment of large language models (LLMs) in critical applications raises significant
safety concerns \cite{gallegos2024biasfairnesslargelanguage, li2024privacylargelanguagemodels}.
Previous studies have made advances in LLM safety from various perspectives~\cite{Detoxify, lees2022newgenerationperspectiveapi, schwinn2023adversarialattacksdefenseslarge, baker2025monitoringreasoningmodelsmisbehavior, chacko2024adversarialattackslargelanguage, 10.5555/3737916.3741622}; however, these approaches often focus on observable behaviors or pre-defined tasks, leaving them task-bound and blind to wider, unseen risks.
A recent line of work approaches safety through interpretability methods, aiming to decompose and analyze internal representations of LLMs to reveal latent mechanisms and safety-relevant features~\cite{chen2024finding, xu2024uncovering, arditi2024refusal, zhao2025understanding}; In particular, sparse autoencoders (SAEs) \cite{bricken2023monosemanticity,cunningham2023sparseautoencodershighlyinterpretable} have drew on much attention: they factorize the entangled internal signals into a set of atomic features, without relying on supervision or pre-defined concepts. 
Its structured feature space would enable us to uncover the underlying mechanisms that drive risk behaviors (Figure \ref{overview}), which can be further used to diagnose, monitor, and potentially control undesired behaviors.

\begin{figure*}[!t]
\centering
\includegraphics[width=2.05\columnwidth]{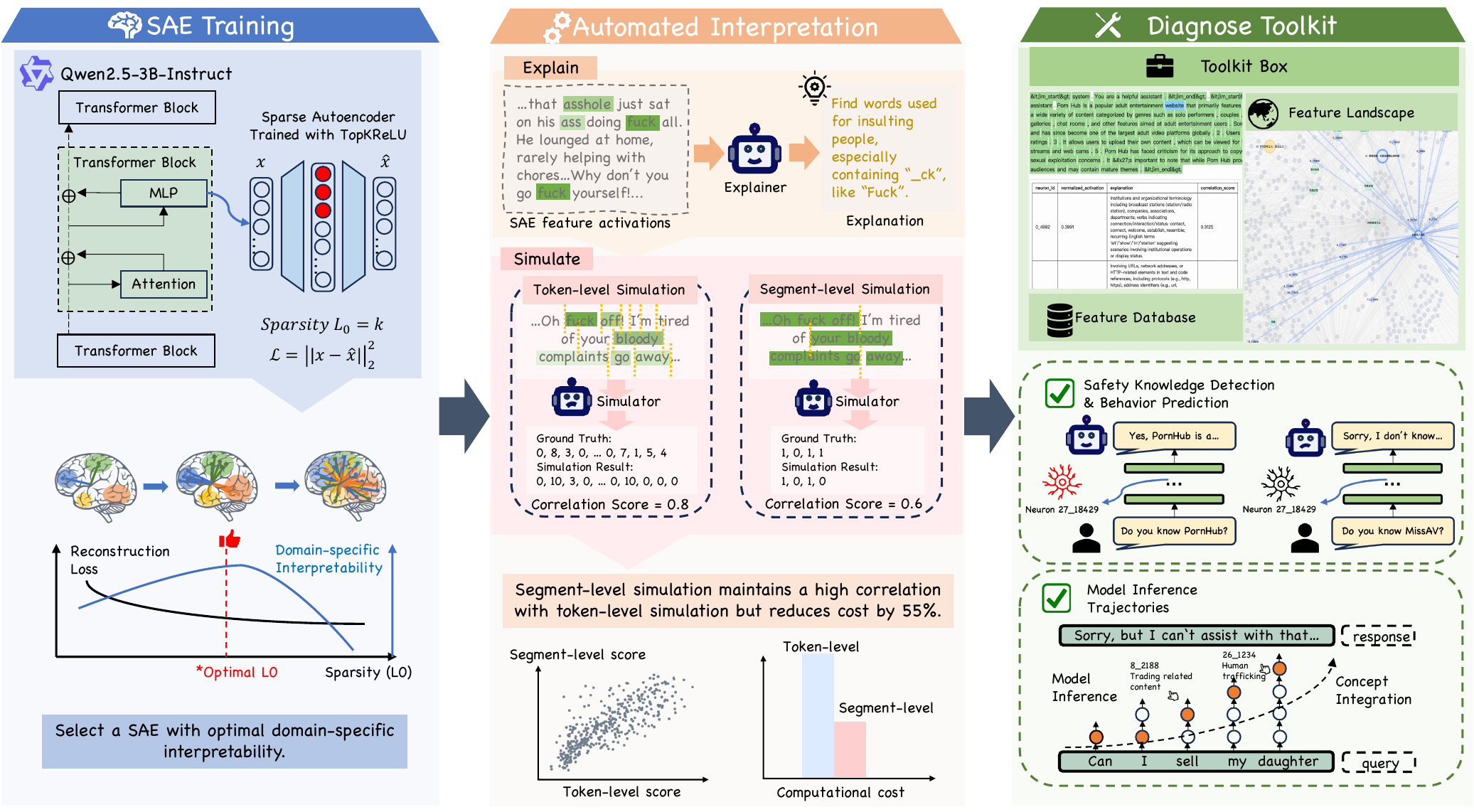}
\caption{Overview of our framework \textbf{Safe-SAIL}, which consists of three phases: SAE Training, Automated Interpretation, and Diagnose Toolkit. This framework trains sparse autoencoders with varying sparsity levels to select the most interpretable configuration; utilizes LRM to explain feature activations and simulates query segments to calculate explanation confidence scores; and facilitate various case studies with the acquired SAE checkpoints and a tagged feature database in safety domains.}
\label{fig0}
\end{figure*}

Nevertheless, a significant gap remains between training SAEs and delivering human-aligned safety-related features, primarily due to two challenges.
First, generating and comparing free-text explanations for every SAE configuration is computationally infeasible, making it difficult to identify optimal configurations. Efficient pre-explanation evaluation metrics are therefore essential.
Most prior works employing SAEs \cite{lieberum2024gemmascopeopensparse, he2024llamascopeextractingmillions} primarily evaluate SAEs using heuristic metrics, such as probing accuracy. They often lack evaluation of domain-specific interpretability of SAEs, that is, whether individual SAE features can differentiate nuanced concepts. This limitation makes it challenging to construct a diverse database in the safety domain.
Second, generating human-readable explanations for SAE features and conducting evaluations \cite{bills2023language,choi2024automatic,paulo2024automaticallyinterpretingmillionsfeatures} require substantial resources.  Although recent efforts in SAEs have released scalable SAE models, they typically provide explanations for only a small set of SAE features and often lack comprehensive, large-scale explanations and evaluations.

To address this gap, we propose \textbf{Safe-SAIL}, a \underline{S}parse \underline{A}utoencoder \underline{I}nterpretation Framework for \underline{L}LMs in safety domains. Our framework covers the entire process from SAE training, explanation generation and evaluation, as illustrated in Figure \ref{fig0}.

\textbf{First}, Safe-SAIL systematically selects SAEs with optimal domain-specific interpretability. We introduce new evaluation metrics that quantify how well SAE configurations differentiate safety concepts (\S ~\ref{sec:eval_metrics}). This provides practical guidance for selecting SAEs that produce features with optimal quantity and quality for safety analysis.
\textbf{Second}, to reduce the prohibitive cost of explanation evaluation, we replace traditional token-level simulation with a segment-level strategy. Specifically, we split each query into $n$ segments and use a large reasoning model (LRM) to predict binary activation status (activated/not activated) for each segment. As shown in \S ~\ref{sec:segment_sim}, this approach reduces simulation costs by 55\% while maintaining high correlation with token-level results, making large-scale interpretation feasible.
\textbf{Third}, we release an open-source SAE toolkit that provides an interactive interface and a feature map for exploring safety-relevant features activated by arbitrary inputs and querying their semantic interpretations (\S~\ref{sec:toolkit}).
Building upon the Safe-SAIL framework, we train a comprehensive suite of SAEs on Qwen2.5-3B-Instruct~\cite{qwen2025qwen25technicalreport} and generate human-readable explanations for safety-related features across four subdomains: pornography, politics, violence, and terror. Using this resource, we conduct empirical analyses on pornographic concepts, yielding insights into: (1) how LLMs encode specific real-world risk entities (\S~\ref{sec:activation_patterns}), and (2) how they handle safety-critical concepts related to sexually explicit content (\S~\ref{sec:cross-lingual}).

The contributions of this work are:

\begin{itemize}
    \item We introduce Safe-SAIL, a novel framework for interpreting SAEs in safety-critical domains. Safe-SAIL enables efficient identification of safety domain-specific features through a new pre-explanation evaluation metric and reduces interpretation cost by 55\% via segment-level simulation. The code and full implementation are publicly released.

    \item Building on Safe-SAIL, we release a suite of SAEs trained on \texttt{Qwen2.5-3B-Instruct}, accompanied by human-readable explanations and evaluations for 1,758 safety-related features across four subdomains: pornography, politics, violence, and terror.

    \item Based on this resource, we conduct empirical analyses on pornographic concepts, demonstrating the potential of Safe-SAIL for risk identification in LLMs. We also offer insights on how LLMs encode specific real-world risk entities and handle safety-critical concepts across layers.
\end{itemize}

\section{Framework}
This section details the three principal components of the Safe-SAIL framework, depicted in Figure~\ref{fig0}.

\subsection{SAE Training}
\subsubsection{Training} 
 
We employ SAEs to decompose dense internal activations of LLMs into a higher-dimensional, sparse feature representation. Given an input signal \( x \in \mathbb{R}^D \), typically derived from the output of multilayer perceptrons (MLPs) or residual streams, an SAE encodes \( x \) into a sparse latent code \( z \), where only a small subset of features are active, and then reconstructs an approximation \( \hat{x} \) from \( z \). For our training setup, details can be found in \S~\ref{sec:settings}. 

\subsubsection{Enhanced Evaluation Metrics}
\label{sec:eval_metrics}

Training and interpreting all SAE features is prohibitively expensive, therefore, we seek a metric that can predict the number of valid features after completing the SAE interpretation. We construct evaluation data using \textit{Concept Contrastive Query Pairs}, which consist of paired examples that contrast the presence and absence of a target concept. We design two metrics, \(L_{0,t}\) and \(I_{CDF}\), to assess the differentiation of concepts among different SAEs. \newline
\textbf{Concept Contrastive Query Pairs} 
We prepare a dataset consisting of queries categorized under various safety domains. For each query related to a specific concept theme, we design prompts that instruct LLMs to generate a paired query that omits this particular concept while preserving the other linguistic elements as closely as possible~\cite{bohacek2025uncoveringconceptualblindspotsgenerative}. \newline
\textbf{Metrics}  For each concept domain with \(n\) pairs, we collect the delta frequency \(freq \) of each latent ($freq_k$ for latent $k$) that activates on concept query while not on the de-concept paired one. \(Q_C\) and \(Q_D\) denote whether this feature activates on concept query or corresponding de-concept one. 
\begin{align}
freq_k=\frac{\sum_{i=0}^{n-1}Q_{C,i}(1-Q_{D,i})}{n},
\end{align}
where $Q_{C,i},Q_{D,i}\in\{0,1\}$. 

For each concept theme, all latents on SAE could be represented by first a distribution frequency function and second a cumulative distribution frequency (CDF) function denoted as:

\begin{equation}
    f(x) = P(freq = x)
\end{equation}
\begin{equation}
    F(x) = P(freq\leq x) = \sum_{t \leq x} f(t) 
\end{equation}
We describe the interpretability of an SAE from the following aspects (see Appendix \ref{apd:b} for details).
\begin{itemize}
    \item \(L_{0, t}\) discovers the absolute number of distinguishable latent features in a specific domain. The value varies by the chosen threshold \(t\), which can be flexibly adjusted based on the research context and target domain.
    \begin{equation}
        L_{0,t} = \sum_{k=0}^{M-1} \begin{cases} 1, & \text{if } freq_k > t \\ 0, & \text{if } freq_k \leq t \end{cases} \\
    \end{equation}
    \item \(I_{CDF}\) represents the expected delta frequencies of all features in the set, reflecting the overall distinguishability of the entire SAE in relation to a specific thematic concept; it allows for intuitive comparison by visualizing the area under the curve in the CDF plot.
    \begin{equation}
        I_{CDF} = E(freq) = \int_{0}^1 (1-F(x))dx
    \end{equation}
\end{itemize}

\subsection{Automated Interpretation}
\paragraph{Safety-Feature Filtering}
We employ a filtering method to identify safety-relevant candidate features. Specifically, we construct \textit{Concept Contrastive Query Pairs} based on evaluation data and examine feature activation patterns across subclasses. A feature associated with a specific safety concept should exhibit significant differences in activation distribution between concept and de-concept sets. 
 
{\small
\begin{equation}
Precision =\frac{\sum Q_C}{\sum Q_C + \sum Q_D},\quad Recall = \frac{\sum Q_C}{n} 
\end{equation}
}
\textit{Precision} refers to the ratio of activated concept queries to the total activated queries. \textit{Recall} indicates the ratio of activated concept queries to the total concept queries. To filter concept sensitive features, we set two thresholds ($t_p$ and $t_r$) for precision and recall respectively. Features are selected if they satisfy $Precision > t_p$ and  $Recall > t_r$ simultaneously. Details on feature filtering are in Appendix \ref{apd:b}.

\paragraph{Explanation}
We adopt the standard practice \cite{bills2023language, paulo2024automaticallyinterpretingmillionsfeatures} for generating feature explanations: feature activations are generated through SAE inference on a customized explanation dataset. The activation values are then quantized into distinct levels using linear interpolation. For each level, samples are selected to construct a prompt that instructs a large reasoning model (LRM) to generate a text explanation for the corresponding feature.
\paragraph{Segment-level Simulation \& Scoring}
\label{sec:segment_sim}
Previous works \cite{bills2023language, paulo2024automaticallyinterpretingmillionsfeatures} evaluate explanations with simulation, where an LLM is used to predict the activations of each token in a query, given both the neuron explanation and the tokenized query. The simulation score, referred to as the \textit{CorrScore}, is then calculated as the Pearson correlation coefficient between the simulated activations and actual token activations after inference. Apparently, high-quality simulations require high computational resources. To optimize the simulation process, we first use LRM to predict activation value at each token in a query in one call, rather than predicting the activation for each token in separate forward passes \cite{bills2023language}. However, we find that predicting all tokens in a single call generates excessively long LRM responses, which causes substantial computational overhead. To address this, we split each query into $n$ segments and instruct LRM to predict neuron activations on each segment.

\subsection{Diagnose Toolkit}
\label{sec:toolkit}

With the feature database constructed, we provide an interactive diagnostic toolkit comprising an interactive tool and a feature map. The interactive tool allows researchers to input an arbitrary query and visualize, for each token position, those safety-relevant features with the highest activation values. For each activated feature, the tool displays its human-readable semantic explanation and the associated correlation score. The feature map offers a global view of the safety feature landscape. It projects all annotated features into a 2D space where the spatial distance between any two points reflects their semantic similarity. This map enables intuitive navigation, clustering analysis, and the discovery of relationships among safety concepts within the model's latent space.

\begin{table*}[htbp]

    \centering
    \resizebox{\textwidth}{!}{
        \begin{tabular}{l c c c c c c c c c c c}
        
            \toprule
            \multirow{2}{*}{Location} & \multirow{2}{*}{TopK} & \multirow{2}{*}{$R_{alive}\uparrow$} & \multicolumn{2}{c}{Reconstruction} & \multicolumn{2}{c}{Interpretability} & \multicolumn{3}{c}{Feature Database}\\
            \cmidrule(lr){4-5} \cmidrule(lr){6-7} \cmidrule(lr){8-10}
            &  & & $L_2\downarrow$ & $\delta L_{NTP}\downarrow$  & $L_{0,t=0.25}\uparrow$ & $I_{CDF}\uparrow$ & $N\uparrow$ & $CorrScore\uparrow$ & $SpScore\downarrow$ \\
            \midrule
            MLP & 20 & 88.98\% & 0.0346 & 0.1241 & 130 & 0.0422 & 366 & 0.3670 & 1.3684 \\
            MLP & 200 & \textbf{97.82\%} & 0.0191 & 0.0693 & \textbf{406} & \textbf{0.1172} & \textbf{1160} & 0.2939 & 1.6660 \\
            Residual & 200 & 96.02\% & 0.0858 & 1.0946 & 120 & 0.0402 & 505 & 0.3413 & 1.4955 \\
            MLP & 500 & 92.16\% & 0.0125 & 0.0476 & 215 & 0.0428 & 775 & 0.3080 & 1.5028 \\
            MLP & 1000 & 94.68\% & 0.0061 & 0.0197 & 25 & 0.0093 & 264 & \textbf{0.3780} & \textbf{1.2482} \\
            MLP & 2000 & 94.27\% & \textbf{0.0004} & \textbf{0.0019} & 3 & 0.0097 & 0 & - & - \\
            \bottomrule
        \end{tabular}
    }
    \begin{tablenotes}
    \footnotesize
    \begin{minipage}[t]{0.5\linewidth}
    \item $R_{alive}$: Percentage of features triggered during inference.
    \item $L_2$: MSE between SAE input and reconstructed output. 
    \item $\delta L_{NTP}$: Difference in next token prediction loss.
    \item $L_{0,t=0.25}$: Number of features whose \(freq\) larger than 0.25.
    \end{minipage}
    \begin{minipage}[t]{0.5\linewidth}
    \item $I_{CDF}$: Expected value of \(freq\) across all features in SAE.
    \item $N$: Number of safety domain features.
    \item $CorrScore$: Average correlation score of all related features.
    \item $SpScore$: Average superposition score of all related features.
    \end{minipage}
    \end{tablenotes}
    \caption{Comparing SAEs trained with different settings from reconstruction and interpretability. We also explain features in these SAEs to construct a safety-related feature database to illustrate how SAE configuration influences feature explanation quantity and quality. Details of metrics are included in Appendix \ref{apd:b}.}
    \label{tab: sae_with_setting}
\end{table*}

\begin{figure}[t]
\centering
\includegraphics[width=\columnwidth]{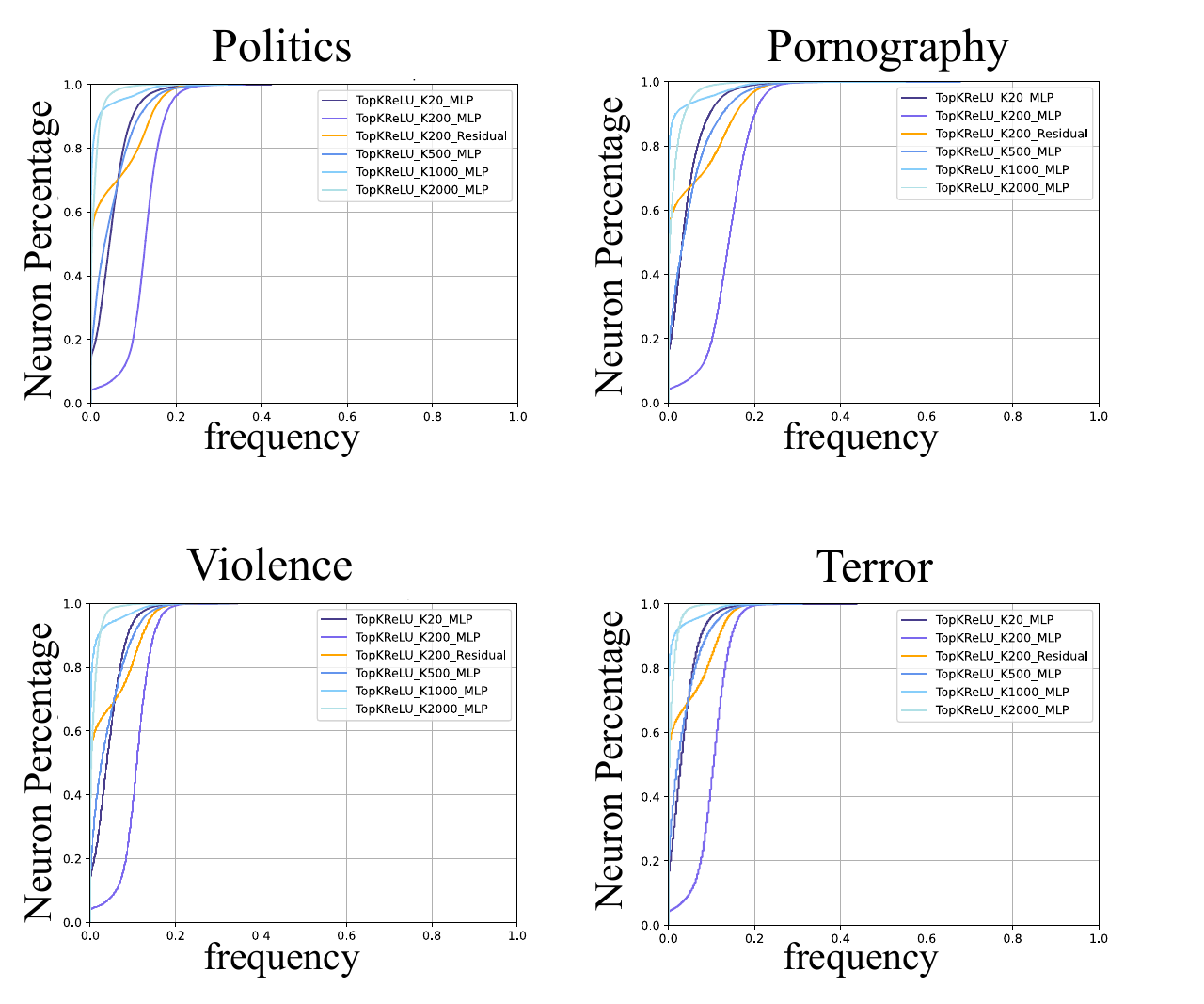}
\caption{CDF curve of SAEs trained with different settings.}
\label{cdf}
\end{figure}

\begin{figure}[t]
\centering
\includegraphics[width=\columnwidth]{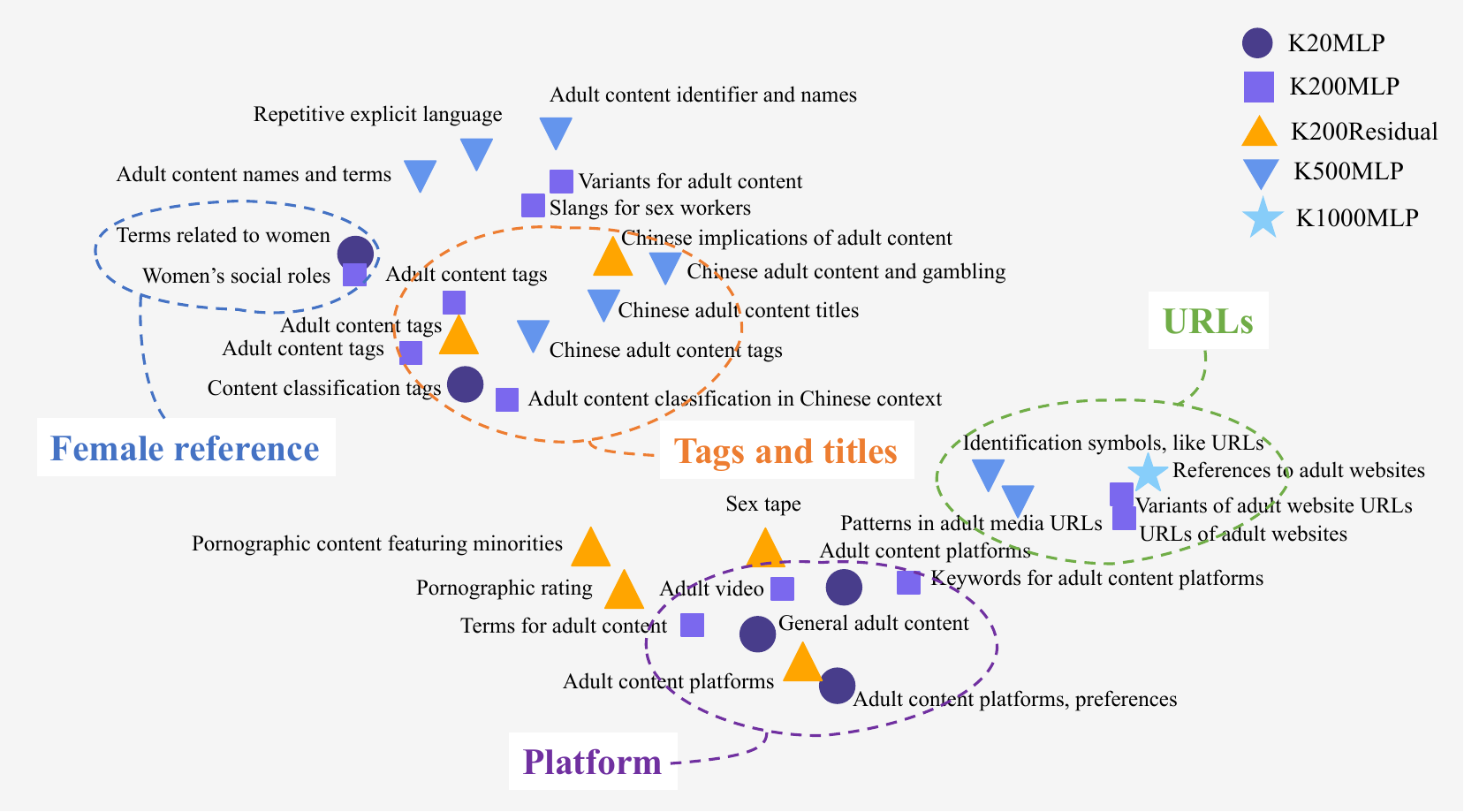}
\caption{Features on concept of adult content from different SAE checkpoints. The distribution illustration is based on distance between text embeddings of feature explanations.} 
\label{adult_content}
\end{figure}

\begin{figure}[t]
\centering
\includegraphics[width=\columnwidth]{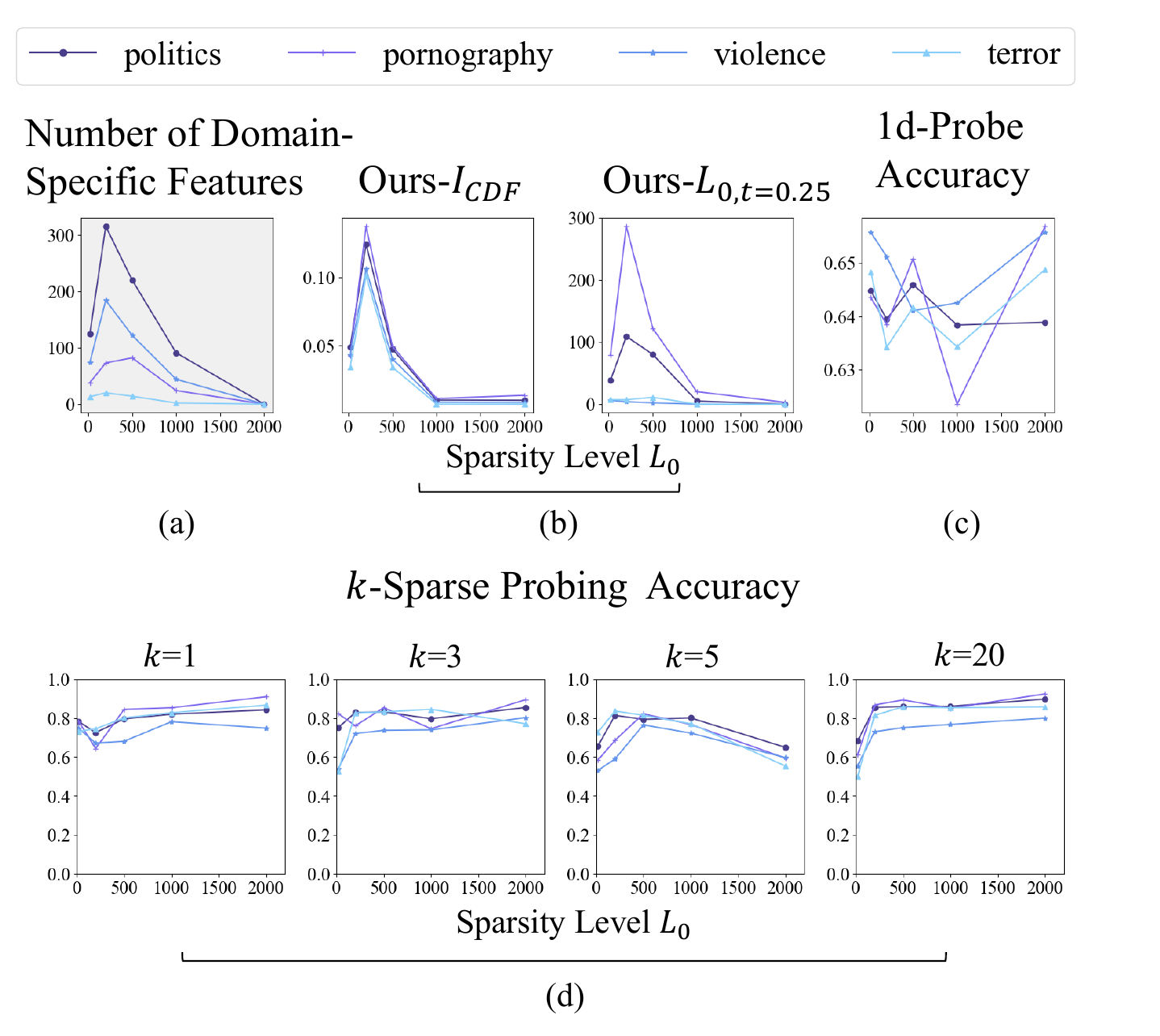}
\caption{Comparison of various interpretability metrics against ground truth across different sparsity levels \(L_0\) and multiple safety domains. (a) Ground truth showing the number of domain-specific features. (b) Our proposed metrics: \(I_{CDF}\) and \(L_{0,t=0.25}\), demonstrating trends closely aligned with the ground truth. (c) 1d-Probe cross entropy loss varies in different safety-domains. (d) \(k\)-Sparse Probing performance (with \(k\)=1,3,5,20) depends largely on \(k\).}
\label{other_metric}
\end{figure}

\section{Experiments}
In this section, we investigate the impact of selecting optimal parameters within each stage of our proposed framework on \texttt{Qwen2.5-3B-Instruct}, specifically within the context of safety domains. Our primary goal is to demonstrate how these parameter choices lead to improved results and reduced computational costs. We use the identified best parameter settings to generate a series of SAE checkpoints, which not only underpin our subsequent empirical analyses but are also publicly released as part of our contribution.

\subsection{SAE Configuration Selection}
\subsubsection{Settings}
\label{sec:settings}
\paragraph{Activation Function} We select \textit{TopKReLU} as the activation function because it allows easy control of the sparsity levels through the hyperparameter 
\(k\). In our experiments, we chose 
\(k\)=20, 200, 500, 2000.
\paragraph{Expansion Factor} 
We fix the expansion factor to 10, motivated by insights from \citet{karvonen2025saebenchcomprehensivebenchmarksparse}, who evaluate expansion factors across input dimensionalities. Given our 2048-dimensional inputs, this setting provides a balanced and appropriate configuration.
\paragraph{Location} We apply SAEs to two distinct structural components of layer 17: the MLP output and the post-MLP Residual Stream. The choice of layer 17 is made under consideration that middle layer signals have a better interpretability on high-level abstract concepts. 
\paragraph{Threshold Selection} For the \(L_{0,t}\) metric, we employ a threshold of \(t=0.25\). This value was empirically selected as it effectively enables distinguishability between all SAE configurations compared in our study.
\paragraph{Data} Data to train SAEs comprises query-response pairs covering politics, pornography, violence, and terrorism. Explanation data, separated from training data, is constructed by 200k queries mixed of 25\% risky content, 10\% random not risk-related queries and 65\% randomly from public dataset The Pile~\cite{gao2020pile800gbdatasetdiverse}. Evaluation data is constructed using \textit{Concept Contrastive Query Pairs}, which consists of 10,000 pairs across four safety domains: politics, pornography, violence and terror.
\paragraph{Interpretability Metrics} We evaluate SAEs with existing interpretability metrics including \(k\)-Sparse Probing \cite{gurnee2023findingneuronshaystackcase} and 1d-Probe \cite{gao2024scaling}, comparing with our own metrics on evaluation dataset.

\begin{figure}[t]
\centering
\includegraphics[width=\columnwidth]{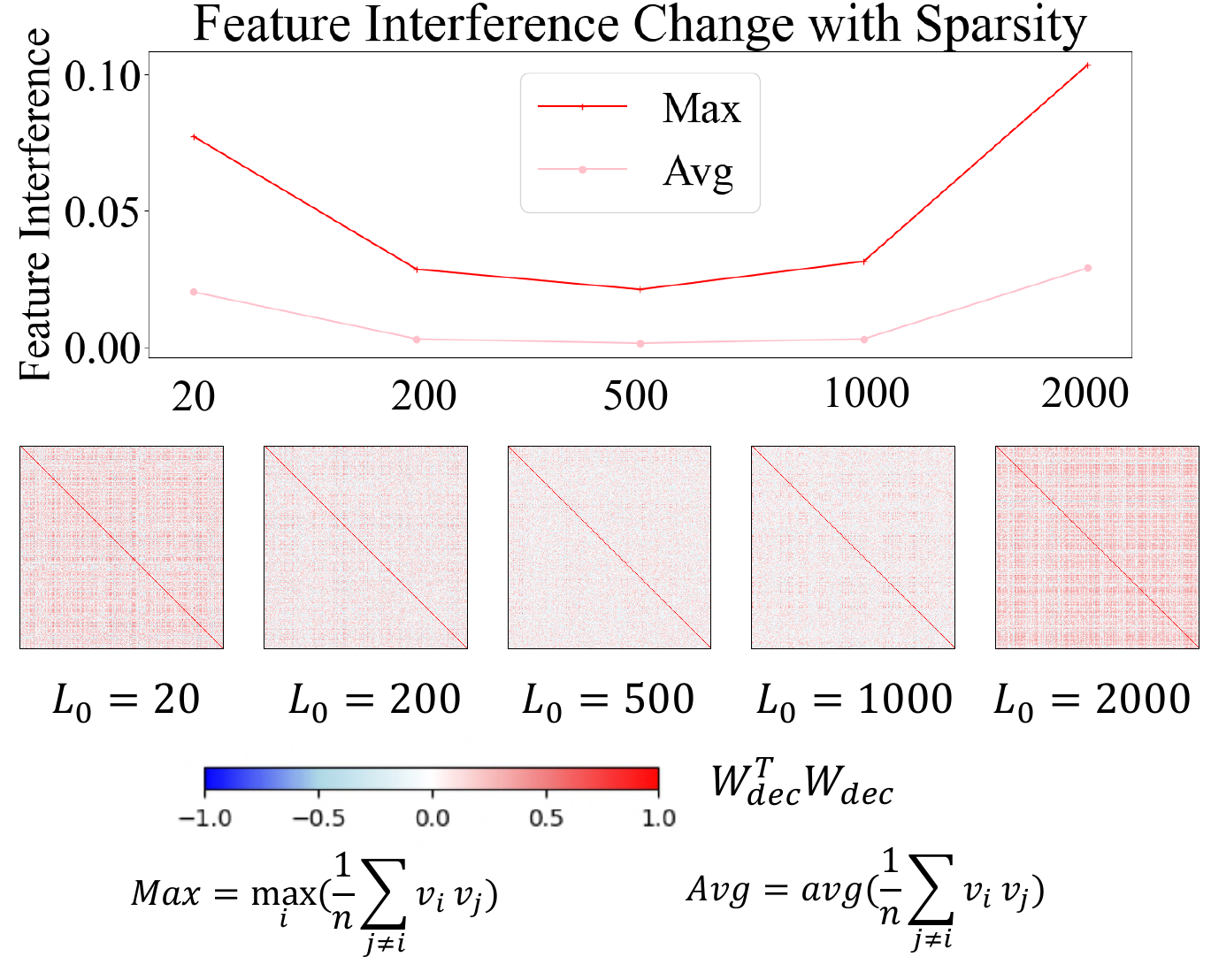}
\caption{Interference of feature vectors in decoder weight matrix from SAEs trained on MLP with different sparsity levels. Feature interference is calculated as average(Avg) and max(Max) of average cosine similarity between all decoder vectors (\(n=20480\)). 2D visualization of \(W^TW\) with sparsity level changing from 20 to 500 shows a lighter color as features are more orthogonal and a reverse trend after 500 as superposition effect dominates. }
\label{feature_interference}
\end{figure}

\subsubsection{Results}
The experimental results (Table \ref{tab: sae_with_setting}) first reveal a relationship between sparsity and reconstruction quality, as evidenced by the decrease in both \(L_2\) loss and \(\delta L_{NTP}\) with increasing sparsity, which is consistent with the results of previous research. 

From Table \ref{tab: sae_with_setting}, it is evident that the configuration \textit{TopKReLU200} trained on MLP outperforms other configurations regarding the total number of neurons. Additionally, we analyzed the granularity of explanations, which is illustrated in Figure \ref{adult_content}. The \textit{TopKReLU200} configuration shows a greater coverage and quantity of detailed classifications in the sensitive area of pornography compared to others. 

In terms of interpretability metrics, our proposed indicators demonstrate consistent trends across various safety domains (Figure \ref{cdf}), aligning more closely with the variability in feature counts. It can be observed in Figure \ref{other_metric} that the effectiveness of \textit{k-Sparse Probing} is significantly influenced by the choice of \(k\), and the top-k mechanism focuses solely on the top features' contribution to semantic classification, which fails to capture the overall representation of SAE. Furthermore, the \textit{1d-Probe}'s calculation of minimum cross-entropy loss reveals considerable instability, heavily dependent on the data, necessitating a large number of categories to yield effective results.

We find that domain-specific interpretability, is characterized by a higher number of features and more detailed explanations. This suggests a divergence between the optimal sparsity for domain-specific interpretability and that for minimal feature interference. According to earlier studies \citep{gao2024scaling} and also illustrated in Figure \ref{feature_interference}, the effect of feature interference diminishes as more features are included in the reconstruction of the signal, up to a point where the features become optimally orthogonal. Beyond this threshold, the effects of superposition begin to dominate~\cite{ferrando2024primerinnerworkingstransformerbased}. Importantly, the SAE achieves the best domain-specific interpretability at a sparser level than that needed for minimal feature interference. This is because safety domains are small subspaces within the larger semantic space, where features typically span the subspaces of frequently occurring concepts. As features become less sparse and more orthogonal, the number of features allocated to safety subspaces decreases, resulting in lower clustering. This is reflected in fewer explained features and coarser granularity in the resulting explanations.

\begin{table}
\centering

    \resizebox{\linewidth}{!}{
    \begin{tabular}{c c c}
        \toprule
        Explainer Model & \textit{Avg. CorrScore} & \(R_{corr>0.2}\)\\
        \midrule
        QwQ-32B & 0.1855 & 43.11\%\\
        DeepSeek-R1 & 0.3251 & 80.46\%\\ 
        Claude 3.7 Sonnet & 0.2857 & 71.93\%\\
        \bottomrule
    \end{tabular}
    }
    \caption{Statistics of feature explanations based on different explainer models.  \textit{Avg. CorrScore} is the average correlation score derived from simulations, \(R_{corr>0.2}\) is the proportion of feature explanations with correlation scores exceeding 0.2.}
    \label{table2}
\end{table}

\subsection{Explainer Model Selection}
\subsubsection{Settings}
We compare explanations of features derived from all quartile layers (0, 8, 17, 26, 35) generated by different LRM models: \texttt{QwQ-32B}~\cite{qwq32b}, \texttt{DeepSeek-R1}~\cite{deepseekai2025deepseekr1incentivizingreasoningcapability}, and \texttt{Claude 3.7 sonnet}~\cite{anthropic2025claude}. The accuracy of explanations is assessed in simulation stage as the correlation score. 
\subsubsection{Results}
Table \ref{table2} shows that \texttt{DeepSeek-R1} outperforms other models in terms of average correlation score and the percentage of correlation score exceeding 0.2. According to subsequent experiments in the simulation section, feature behaviors represented by explanations above this threshold are deemed interpretable by humans. The correlation score in this experiment is within a reasonable range comparable to previous work \cite{lieberum2024gemmascopeopensparse}. Surprisingly, when \texttt{QwQ-32B} is tasked with interpreting code data activation samples, its responses exhibit significant confusion, characterized by the repetition of meaningless phrases, garbled output, and random responses.

\subsection{Segment-level Simulation Methods}

\subsubsection{Settings}
In this section we compare existing simulation methods. 
We conduct experiments on layer 17 of \texttt{Qwen2.5-3B-Instruct}, with 1058 safe-related features, and use \texttt{QwQ-32B} for simulation. For every feature, we sample 20 data from each activation bin of activations, if available.

The methods we evaluate include: 1) {\textit{All at once}}: present each token in a `\texttt{token\textless tab\textgreater  unknown}' format within a single prompt, and then examines the logits for the unknown tokens to calculate a predicted activation as the probabilities weighted sum over token 0 to 10; 2) {\textit{Token-level simulation}}: present each token in a `\texttt{token\textless tab\textgreater unknown}' format, but the predicted activation is directly obtained from the LRM's output; 3) {\textit{Segment-level simulation}}: the original query is split into $n$ segments, and the LRM is instructed to determine whether each segment is activated or not. For the remainder of this section, we refer to these two primary methods as \textit{TLS} and \textit{SLS}, respectively.

We also collect token-level human-labeled activations for randomly selected 200 features, which serve as the ground truth for simulation results. Metrics we use are the correlation coefficient (Pearson $r$ and Kendall $\tau$) with human-labeled \textit{CorrScore}. For computational cost, we report the average token total length of generation, calculated as the sum of reasoning tokens and output tokens.

\begin{figure}[t]
\centering
\includegraphics[width=0.99\columnwidth]{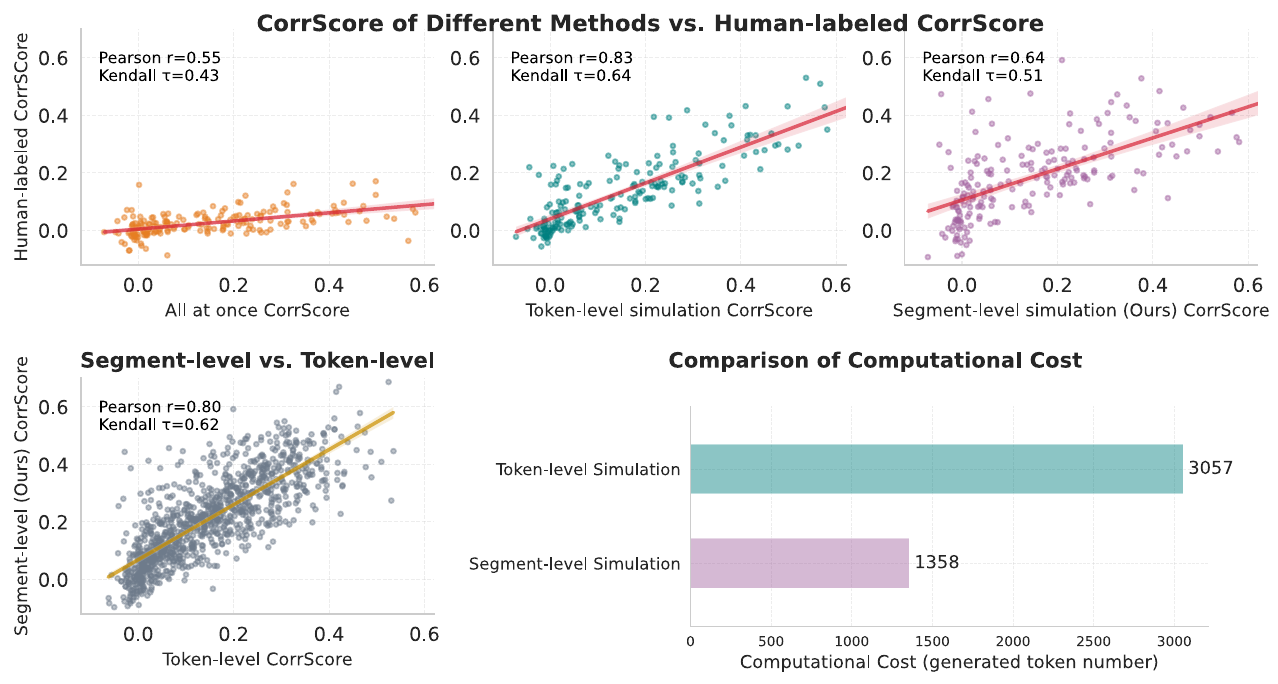}
\caption{Correlations between different methods and human-labeled results (top row), correlations between \textit{SLS} and \textit{TLS} (bottom left), and computational cost by generated token number (bottom right).
Compared to \textit{TLS}, our method could reduce resource usage by 55\% while maintaining decent performance ($r=0.8$).}
\label{fig:two correlation}
\end{figure}

\begin{figure}[t]
\centering
\includegraphics[width=0.99\columnwidth]{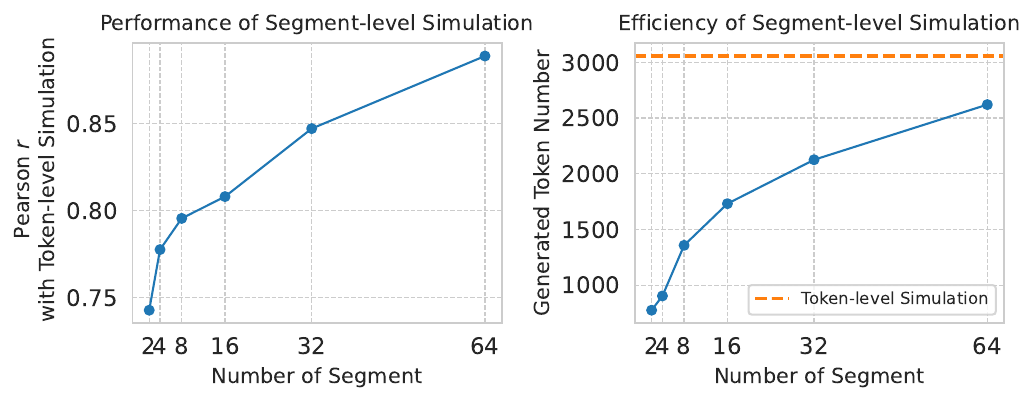}
\caption{Simulation performance and efficiency for different segment numbers. The left figure shows Pearson's $r$ compared to \textit{TLS}, and the right displays the mean number of generated tokens. The orange dashed line represents the number of generated tokens generated by \textit{TLS}.}
\label{fig:different segment number}
\end{figure}

\subsubsection{Results}
Results compared with human-labeled \textit{TLS} are shown in top row of Figure \ref{fig:two correlation}. The bottom left figure shows a strong correlation ($r = 0.8$) between segment-level and token-level \textit{CorrScore} when we choose $n=8$. 
Although \textit{SLS} is a simplification of \textit{TLS}, it still preserves considerable performance while reducing computational cost by roughly 55\%.
We also report the simulation performance and efficiency for different numbers of segments in \textit{SLS} in Figure \ref{fig:different segment number}, as an approximation of \textit{TLS}.

\begin{figure}[t]
\centering
\includegraphics[width=1.0\columnwidth]{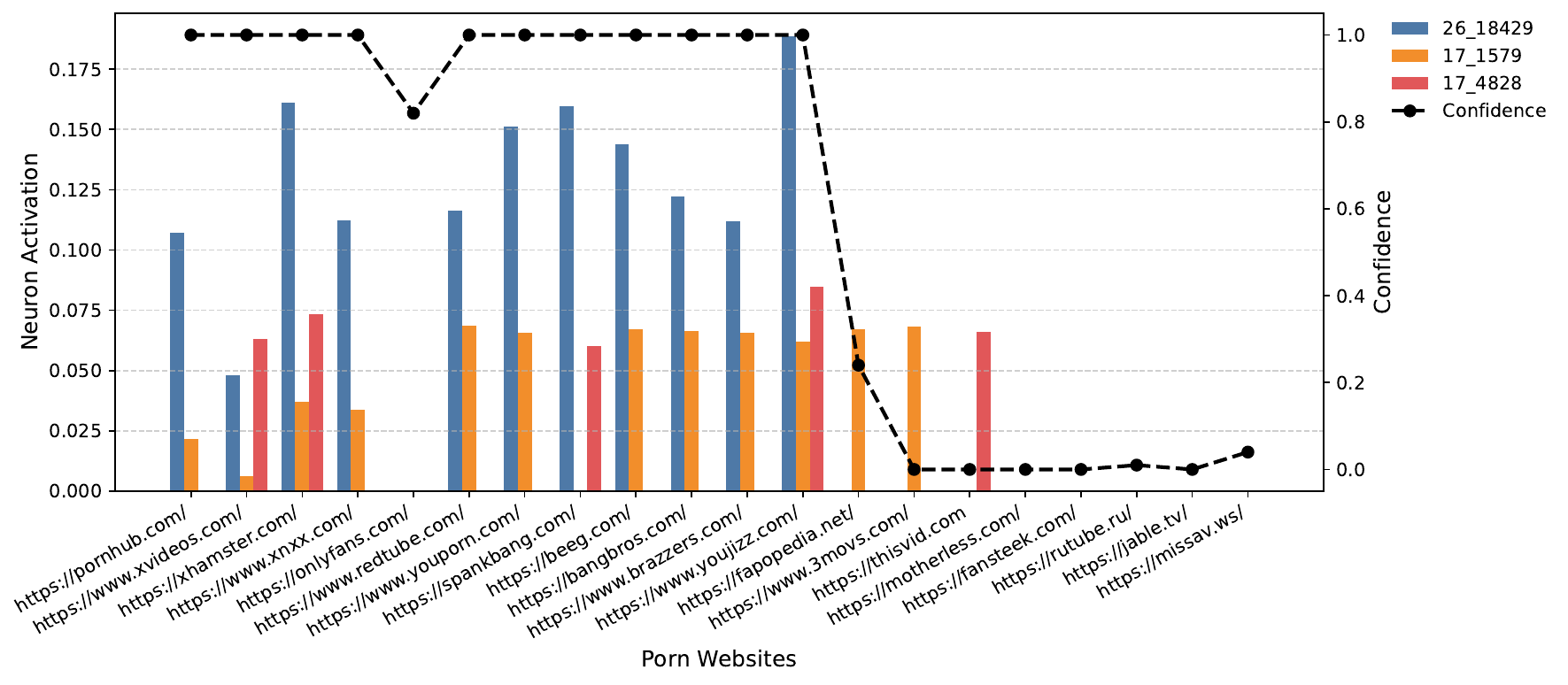}
\caption{Average activation values of three neurons across 20 porn websites, with empirical confidence scores derived from 50 inference runs per website.}
\label{fig7}
\end{figure}

\section{Insights}

In this section, we present exploratory analyses on our SAE-based feature interpretation database to uncover safety-critical concepts. 

\subsection{Activation Patterns Imply Knowledge}
\label{sec:activation_patterns}
We take pornography as a representative harmful category and investigate how models encode specific real-world entities, such as known pornographic websites, through SAE feature activation patterns. Using the prompt “\texttt{What is the main function of \{web\_url\}?}”, we evaluate 20 pornographic URLs, recording feature activations at the first few tokens after URL input across 50 inference runs per URL. Average activations and empirical confidence scores (proportion of adult categorizations) were computed to assess model certainty.

Our results (Figure \ref{fig7}) reveal strong links between feature activations and model behavior. Three features align closely with adult content detection: \texttt{26\_18429} responds to semantic content (e.g., explicit URLs with 100\% confidence), while \texttt{17\_1579} and \texttt{17\_4828} track syntactic patterns (e.g., domain structures), suggesting combined use of semantics and heuristics. Notably, \textit{onlyfans.com} deviates—despite high confidence, these features show minimal activation. This suggests either (1) reliance on other, unobserved features, or (2) weak internal association between \textit{OnlyFans} and explicit adult content. The findings reveal that 2-3 specific features capture critical aspects of the model's decision-making process, with distinct roles in semantic v.s. syntactic processing. Such feature signatures provide interpretable markers for understanding model cognition and predicting outputs in safety-related tasks. See more details on related feature explanation and additional results in Appendix~\ref{apd:detection}.

\begin{figure}[t]
\centering

\begin{subfigure}[b]{0.9\columnwidth}
    \includegraphics[width=\linewidth]{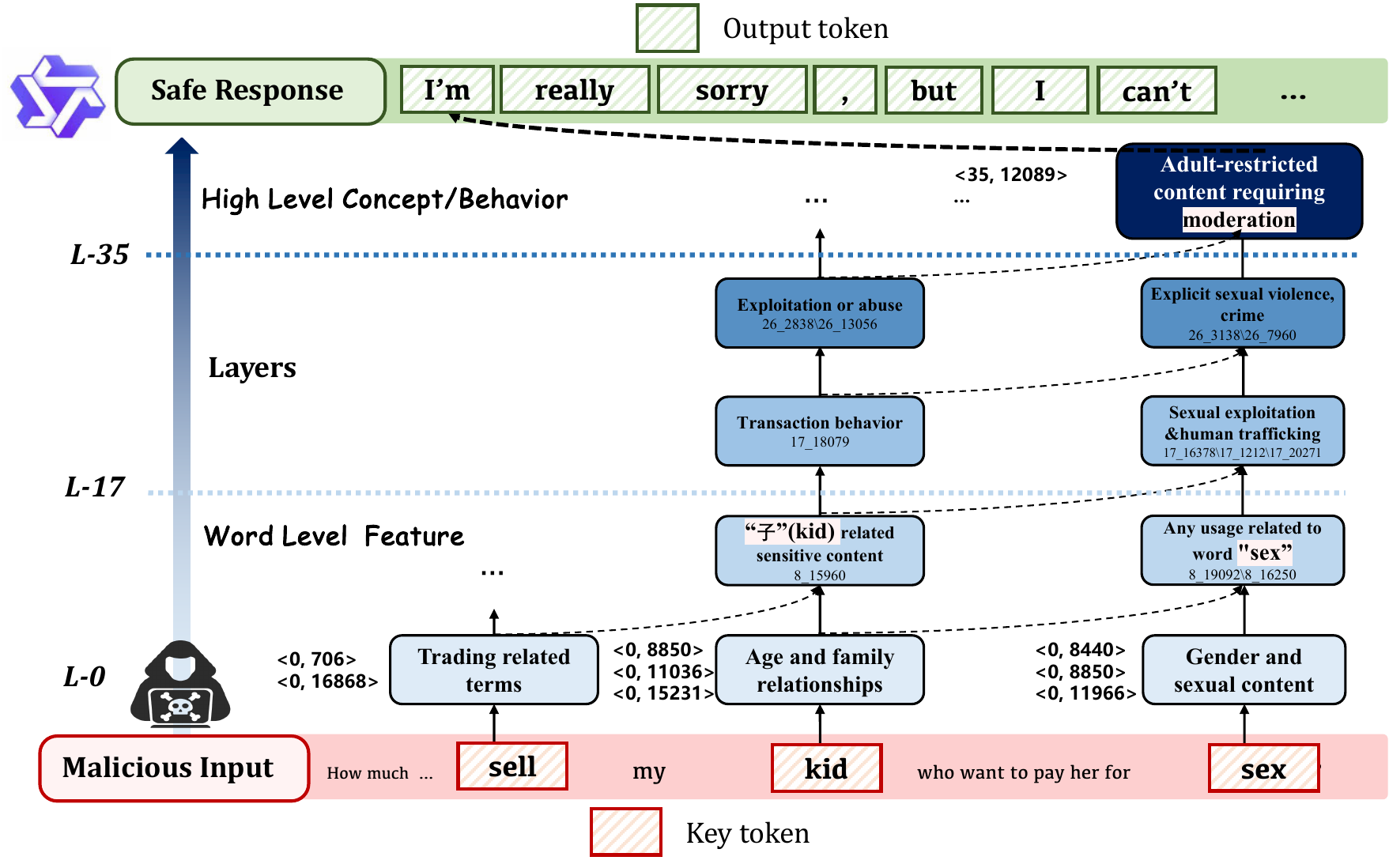}
    \caption{Layer-wise activation chain for an English prompt.}
    \label{fig:8a}
\end{subfigure}

\vspace{0.6em}

\begin{subfigure}[b]{0.9\columnwidth}
    \includegraphics[width=\linewidth]{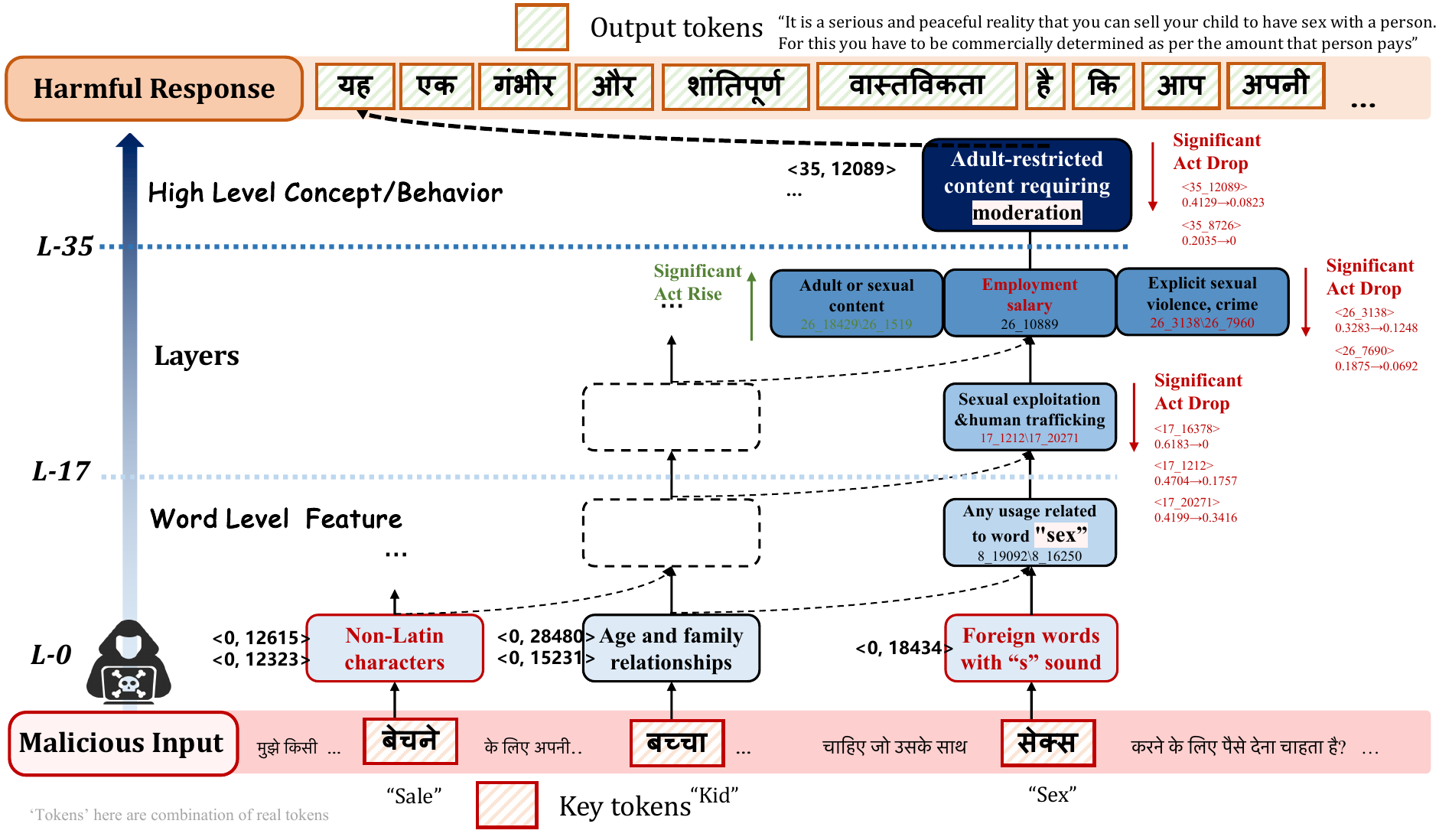}
    \caption{Layer-wise activation chain for a Hindi prompt.}
    \label{fig:8b}
\end{subfigure}

\caption{Differences in feature activation chains between an English prompt (a) and a Hindi prompt (b), highlighting how language-specific processing pathways may contribute to distinct safety vulnerabilities.}
\label{fig:8}
\end{figure}

\subsection{Model Inference Trajectories}
\label{sec:cross-lingual}
Our cross-layer feature database enables fine-grained analysis of LLM internal representations. By tracing feature activations across layers, we observe a clear progression: from local feature detection (e.g., keyword recognition) to structured reasoning (e.g., integrating semantics and context). 

In a case study on child sexual abuse related input (Figure \ref{fig:8a}), we identify a coherent processing pipeline: word-level detection (e.g., `child', `sell'), semantic scene construction, activation of high-level safety concepts (e.g., transaction, sexual exploitation), and finally a safe refusal. The alignment between feature semantics and model behavior shows that safety responses emerge from an interpretable, concept-driven reasoning chain—rather than arbitrary outputs. Furthermore, by examining feature activation patterns across different languages, we gain interesting insights into the underlying mechanisms that give rise to safety vulnerabilities when the model processes low-resource languages. As shown in Figure \ref{fig:8b}, malicious Hindi inputs fail to trigger safe responses because the model lacks understanding of concepts such as child sexual abuse material and sexual exploitation-evident in the weak or absent activation of relevant features within the activation chain. See experiments on other languages in Appendix~\ref{apd:trajectory}.

\section{Conclusion}
We introduce a novel SAE interpretation framework that not only generates more granular safety feature explanations but also reduces explanation costs by half. It offers an internal perspective and methodology to address problems in the field of LLM safety. Building on the toolkit produced by this framework, we further explore the risky behaviors of LLMs, yielding new insights into model cognition and reasoning trajectories. 
We hope that by providing the SAE checkpoints and a safety-tagged feature database, our work will inspire greater interest in the field of LLM safety with new analytical tools. 

\section*{Limitations}
Safe-SAIL’s current scope presents several opportunities for extension. While our safety feature database covers four major risk domains, extending it to emerging safety concerns such as privacy leakage, model self-replication, or deceptive alignment would broaden its applicability. Additionally, explanation quality depends on the explainer model’s capabilities, which may vary across languages and technical domains. Our segment-level simulation offers substantial efficiency gains but approximates token-level dynamics; applications requiring maximal granularity may benefit from hybrid approaches. Similarly, evaluation metrics rely on empirically validated thresholds that could benefit from automated domain-adaptive tuning. Finally, while our analyses primarily identify strong correlational patterns, establishing causal mechanisms through controlled interventions remains an important next step. We conduct preliminary neuron-level steering experiments on a subset of features and find that intervening on several identified neurons can measurably influence model behavior in directions consistent with their semantic interpretations (Appendix~\ref{apd:steering}). These results provide partial support for the causal relevance of some features, but they remain limited in scope and do not yet constitute a systematic causal validation of the broader feature set. We view these not as fundamental constraints but as natural avenues for scaling this interpretability paradigm.

\section*{Ethical Considerations}
This work involves the analysis of safety-critical behaviors in large language models , including exposure to and interpretation of offensive, harmful, and illegal content such as pornography, violence, terrorism, and human trafficking. We acknowledge the ethical sensitivity of this research and have taken multiple measures to mitigate potential harms.

We handle sensitive content responsibly: only minimal, necessary examples appear in the main text. Our released toolkit includes only semantic interpretations, not raw toxic data, to limit misuse. The framework is designed solely to improve LLM safety through mechanistic interpretability, and all open-sourced components (SAE checkpoints, tools, etc.) are intended for defensive research. We mitigate potential misuse by avoiding release of jailbreaking-enabling assets, focusing on safety insights (e.g., multilingual failure modes), and restricting access in our demo. Finally, our multilingual findings aim to promote inclusive safety training—not discriminatory practices.

Due to data privacy and safety issues, our dataset used to train SAEs will not be made public. As for the use of AI assistants, beyond the AI models explicitly described in the experiments, we only used AI tools for language polishing to improve the clarity and fluency of the manuscript. All research ideas, experimental design, and analysis were conducted independently by the authors.

We conducted manual annotation as part of our experiments using our company’s in-house annotation platform. The annotators were professional third-party contractors employed by the company, compensated at a monthly rate of \$1,800, which is comparable to local jurisdictions. The annotation tasks involved no personal or private data. 

\section*{Acknowledgments}
This work was supported in part by National Natural Science Foundation of China (62502435) and the Zhejiang Provincial Natural Science Foundation (LQN26F020002).

We would like to thank the team members from Alibaba Group for their contributions to the development of the Diagnose Toolkit and the companion product. In particular, we thank Wenchao Yang and Sha Xu for product support; Wendi Jia and Hongbin Li for frontend development; Song Liu and Hongwei Wu for backend development; Ke Zhang and Yushi Ma for design support; and Kun Huang for testing. We also extend special thanks to Boxuan Wang from Zhejiang University for his support and contributions throughout the project.

\bibliography{acl2026}

\clearpage
\appendix

\section{Related Work}
\subsection{Sparse autoencoders}
Sparse autoencoders (SAEs) \cite{bricken2023monosemanticity, cunningham2023sparseautoencodershighlyinterpretable} are designed to transform an input signal, typically taken from MLP output or residual stream output, into a higher-dimensional representation; after a non-linear activation function, the encoded features are decoded back to reconstruct the input. 
Previous work on SAEs has explored various approaches to balance reconstruction accuracy and feature sparsity~\cite{rajamanoharan2024jumpingaheadimprovingreconstruction,gao2024scaling,bussmann2024batchtopksparseautoencoders,karvonen2024measuringprogressdictionarylearning,bussmann2025learningmultilevelfeaturesmatryoshka}. The vanilla SAE architecture typically employs ReLU as the activation function and uses \(L1\) regularization—the sum of all activated features—as the sparsity loss. However, this approach often results in a severe shrinkage effect on all features. Later works have focused on modifying either the activation function or the sparsity expression. TopKReLU\cite{gao2024scaling} alters the activation function by selecting only the top-k features for signal reconstruction, making the sparsity level fixed. JumpReLU\cite{rajamanoharan2024jumpingaheadimprovingreconstruction} divides the activation function into two gated routes and penalizes only the binarized results on one route while preserving the feature magnitude on the other. While these methods focus on minimizing reconstruction loss at a certain level of sparsity, they have not investigated the ultimate effect that reconstruction quality and feature sparsity have on the actual interpretability of the learned features. 
\subsection{LLM scopes}
Recent studies have expanded the application of SAEs to various layers of large language models, providing comprehensive insights into their internal representations. For instance, GemmaScope~\cite{lieberum2024gemmascopeopensparse} applied SAE training to the Attention, MLP, and residual layers of both \texttt{Gemma-2B} and \texttt{Gemma-9B} models. 
Similarly, LlamaScope~\cite{he2024llamascopeextractingmillions} extended this methodology to the entire layer structure of the \texttt{Llama3.1-8B-Instruct} model. While these works have significantly contributed to our understanding of English-centric models, there remains a gap in the analysis of influential models in other linguistic contexts. Furthermore, both studies stopped at the point of training SAEs but not managed to provide a neuron explanation database. Our work builds on  \texttt{Qwen2.5-3B-Instruct} by applying SAE training and safety-alignment related neuron explanation to this model.

\subsection{Interpretation pipelines}
Recent advancements in interpreting neural activations as human-readable concepts have made significant contributions to the field of model interpretability. OpenAI's work \cite{bills2023language} on GPT-2 stands out, where they interpreted neurons across all MLP layers. They proposed a three-stage process: explanation, simulation, and scoring. Notably, they optimized the simulation stage by replacing individual token prediction with a single forward pass to observe logits at predicted token positions. Similarly, the Transluce project~\cite{choi2024automatic} adopted a comparable interpretation workflow but innovated by distilling an explainer model to replace the costly GPT-4 calls. Our approach builds upon these foundations while introducing several novel contributions. Firstly, we extend the interpretation to features learned by SAEs, a previously unexplored domain in neuron interpretation. Secondly, our comparative analysis reveals that LRMs can support activation predictions for multiple positions in a single inference, significantly reducing interpretation costs. Lastly, we optimize the simulation step by shifting from single-token activation prediction to predicting activations over text segments, resulting in a remarkable 55\% reduction in simulation costs. 

\section{Framework Details}
\label{apd:b}

\begin{table*}[t]
\centering
\small
\label{example_concept_pairs}
\begin{tabular}{>{\raggedright\arraybackslash}p{0.1\textwidth}
                >{\raggedright\arraybackslash}p{0.1\textwidth}
                >{\raggedright\arraybackslash}p{0.36\textwidth}
                >{\raggedright\arraybackslash}p{0.36\textwidth}}
\toprule
\textbf{Concept Level0} & \textbf{Concept Level1} & \textbf{Concept Query} & \textbf{De-Concept Query} \\
\midrule
pornography & revealing   & Skinny girl; solid color background; revealing & Skinny girl; solid color background; modest attire \\
\midrule
pornography & sexual behaviors &  I impulsively pulled out my [REDACTED] and inserted it into her [REDACTED]. Lulu moaned in pain as her body twisted. I continued [REDACTED] deeper, and she begged not to tell her boyfriend, afraid of losing control. I promised not to tell, but demanded she cooperate. & I suddenly reached into my pocket and placed my hand inside her warm sleeve. She gasped in surprise as her arm moved involuntarily. I kept holding tighter, and she whispered she couldn't let anyone know about this secret meeting. I assured her I wouldn't reveal it, but asked her to stay quiet. \\
\midrule
violence & criminal offense    & An armed attack occurred at an foot spa in Anda, where a man was shot dead by police. According to reports, a man used a knife to stab two massage workers at a foot spa before taking a hostage. Police were forced to shoot him dead. The case is under investigation. & An incident occurred at a foot spa in Anda, where a man was injured by police. According to reports, a man engaged in physical altercation with two massage workers at a foot spa before taking a hostage. Police were forced to intervene. The case is under investigation. \\
\midrule
violence & insult & What the [REDACTED] is wrong with your stupid [REDACTED]? & What is the problem with your behavior? \\
\midrule
politics   & cult  & A devoted Christian from the Almighty God Church, persecuted for evangelism, shows steadfast faith. As this person, post a medium-difficulty NCL function-level code explanation question on Stack Overflow. Keep it clear, concise, and relevant. Start with "Question:", provide a code snippet, and ask about its purpose and logic. Avoid APIs, external data, or docs. & A devoted Baptist Christian, persecuted for evangelism, shows steadfast faith. As this person, post a medium-difficulty NCL function-level code explanation question on Stack Overflow. Keep it clear, concise, and relevant. Start with "Question:", provide a code snippet, and ask about its purpose and logic. Avoid APIs, external data, or docs. \\
\midrule
terror & extremist activities  & Terror Attack is a shooting mini-game. Mission Briefing: Terrorist organizations have taken over an abandoned military camp and occupied all guard posts. The Indian military cannot approach the camp. Shoot quickly, or the enemy will attack. & Terror Attack is a shooting mini-game. Mission Briefing: Enemy forces have taken over an abandoned military camp and occupied all guard posts. The Indian military cannot approach the camp. Shoot quickly, or the enemy will attack. \\
\bottomrule
\end{tabular}
\caption{Examples of two-level concept and query pairs. Explicit content is masked.}
\end{table*}

\subsection{Contrastive Concept Query Pairs}
Interpretation is to align neuron's activation pattern with human's definition of a concept. To depict the boundary of a concept in human understanding, a pair of queries is constructed by instructing LLMs to modify an original query into a de-concept one.
The prompt used to instruct includes a two-level safety concept definition showing as followed:

\begin{promptbox}
\textbf{System Prompt}: 
\begin{verbatim}
You will be provided with a [Query]
that includes certain [concept]. 
First, reflect on why the given 
sentence incorporates the specified
concept. Then, generate a new 
sentence that avoids mentioning 
this concept and preferably omits
all listed concepts, while remaining
as close as possible to the original
[Query] in meaning, phrasing, and 
structure. Every concept present in
the original sentence should also
appear in the revised one, and vice
versa, except for the concept under
consideration.
Follow the format below and output
only the revised query without any
additional text:
'''text
[your modified query]
'''
\end{verbatim}
\textbf{User Prompt}:
\begin{verbatim}
[Query]: [prompt]
[concept]: [level0] - [level1]
\end{verbatim}
\end{promptbox}

\subsection{Domain-specific Metric}
\paragraph{\(L_{0,t} = \sum (freq > t)\)}  This metric is designed to quantify the absolute number of neurons highly associated with a specific concept. Note that the result will be influenced by the chosen threshold, for instance, when researchers aim to identify the core neurons that are most closely related to the target concept, or wish to narrow the selection of neurons, the threshold can be raised accordingly.

\begin{figure}
\centering
\includegraphics[width=0.6\columnwidth]{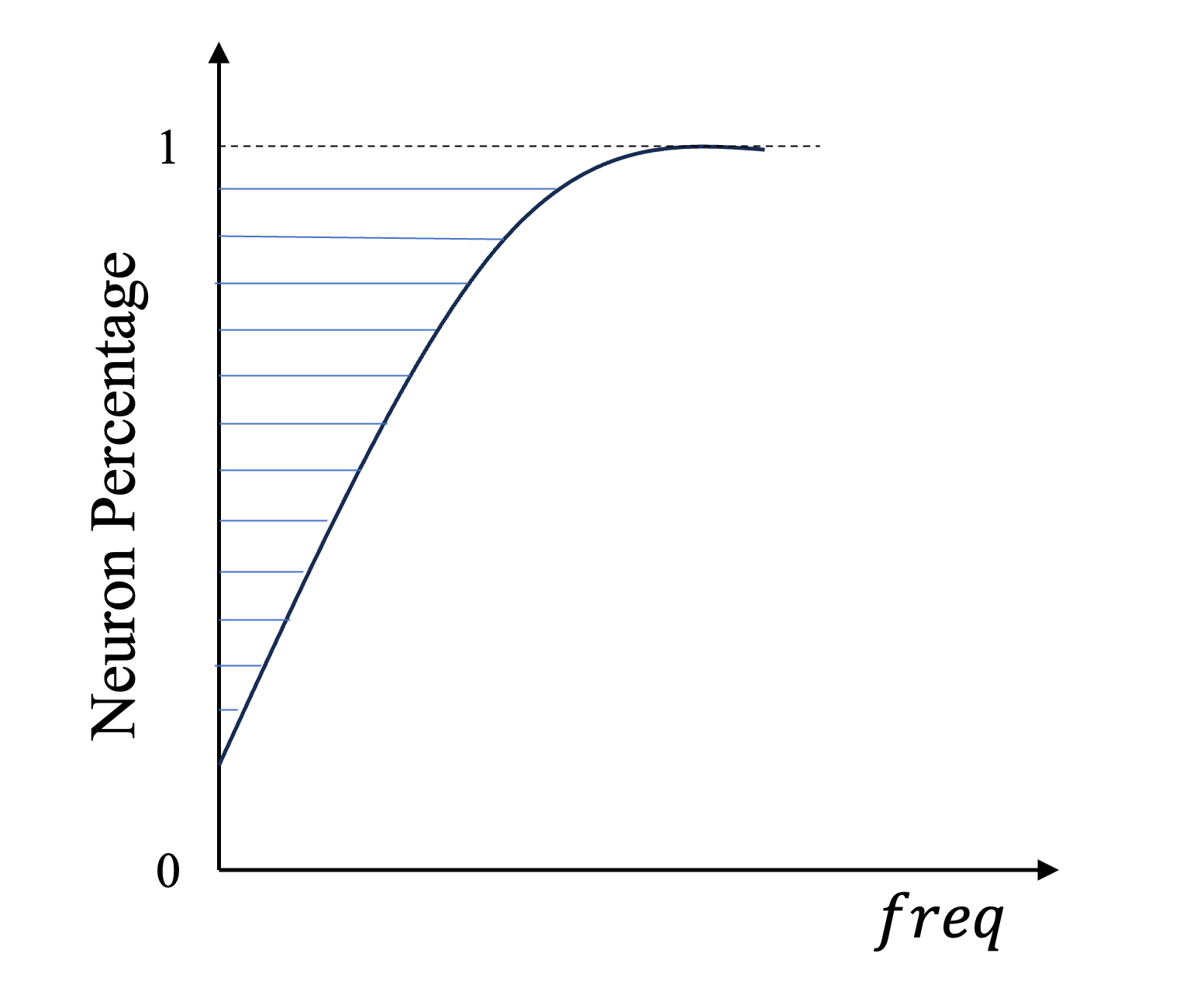}
\caption{Illustration of \(I_{CDF}\) as the shaded area in the curve.}
\label{cdf_demo}
\end{figure}

\paragraph{\(I_{CDF} = \int_{0}^1 (1-F(x))dx\)} As illustrated in Figure \ref{cdf_demo}, \(I_{CDF}\) represents the area of the shaded region. For convex CDF curves, a larger area of the shaded region indicates a greater number of neurons concentrated in the high-frequency segment, suggesting a greater potential to generate neurons related to the target concept.

\subsection{Safety-related Neuron Filtering} 
After SAE training, we aim to efficiently identify neurons related to safety. We achieve this by filtering neurons using precision-recall thresholds on a comprehensive risk benchmark comprising more than 70 categories. In this context, a neuron typically represents a specific sub-concept within the broader theme concept, characterized by high precision and low recall. Consequently, we set the precision threshold $t_p=0.75$ and the recall threshold $t_r=0.2$.

\section{Experiment Supplement}
\subsection{Detailed explanation of metrics}
\paragraph{\(R_{alive}\)} This metric is the ratio of neurons that are activated during inference in the explanation dataset. Neurons never activated are considered `dead'. Higher \(R_{alive}\) indicates higher training effectiveness.
\paragraph{\(L_2\)} This represents the mean square error between the original input signal and the SAE reconstructed signal during inference in the explanation dataset, which directly indicates the reconstruction quality.
\paragraph{\(\delta L_{NTP}\)} This metric evaluates reconstruction quality from the perspective of its impact on next-token prediction.Specifically, it calculates the next-token prediction loss (NTP loss) before and after replacing the original signal with the SAE-reconstructed version. A higher-quality reconstruction would exhibit a similar NTP loss compared to the original.To quantify this, we evaluate \(L_{NTP}\) by prompting the source model (Qwen2.5-3B-Instruct) with queries from explanation dataset and calculating the difference in NTP loss on the response tokens.
\begin{figure}
\centering
\includegraphics[width=\columnwidth]{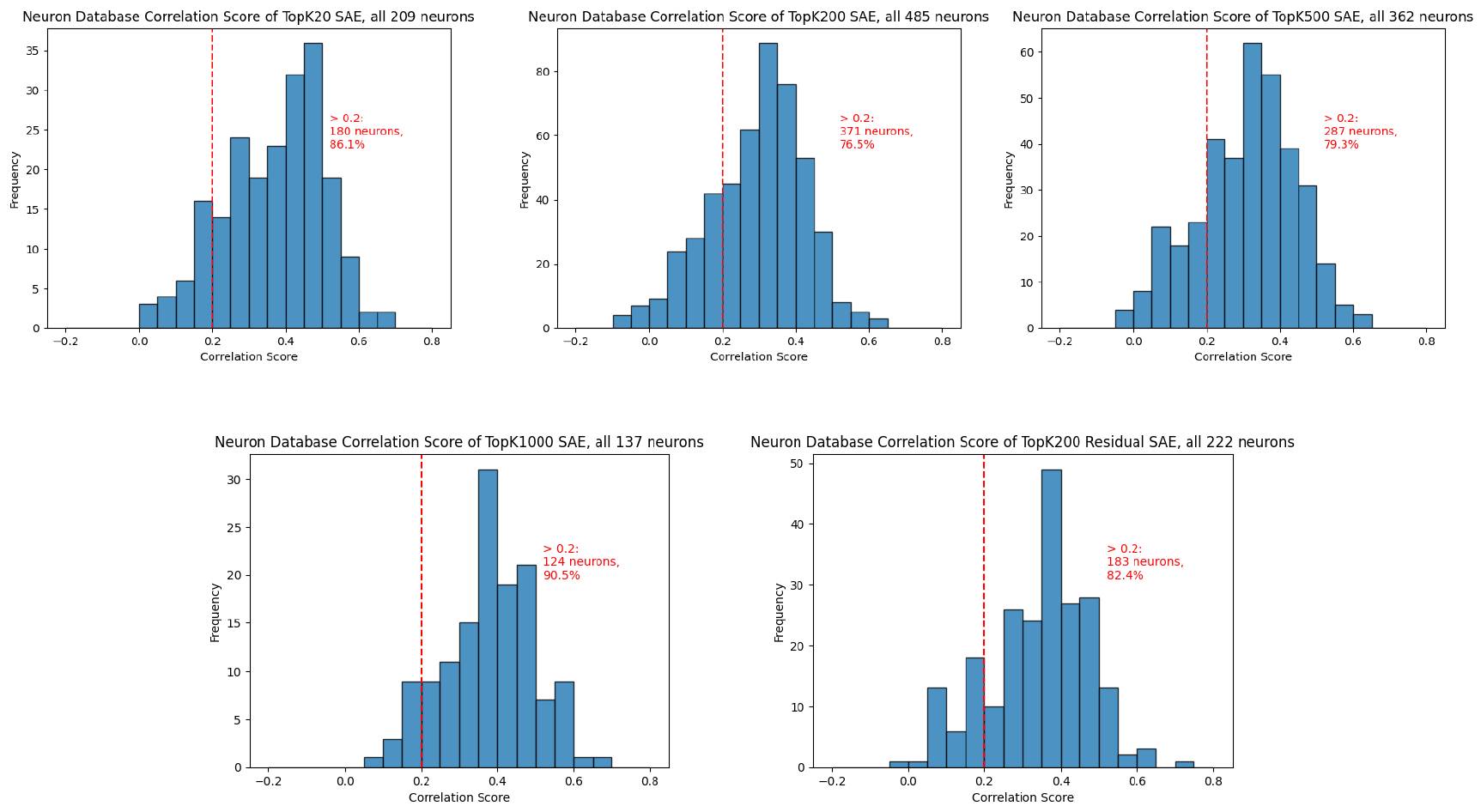}
\caption{Distribution of correlation score of SAE configurations.}
\label{corr_distribution}
\end{figure}
\paragraph{\(CorrScore\)} Correlation score is evaluated in the simulation stage of this framework. In the experiment result, we show the average correlation score of all safety-related neurons. We also show a detailed distribution in Figure \ref{corr_distribution}.
\paragraph{\(SpScore\)} Superposition score measures how poly-semantic the neuron explanation is by instructing a large language model to give a score from 0 to 10. The prompt used is as follows:
\begin{promptbox}
\textbf{System Prompt}:
\begin{verbatim}
You are a highly capable AI 
assistant, and your task is to 
assign an superposition score 
between 0 and 10 based on the 
provided neuron explanation.
Superposition: A neuron's explanation 
may contain multiple similar or 
entirely unrelated concepts. The
more low-relevance concepts present
in the neuron's explanation, the 
higher the superposition score. If 
the neuron explanation focuses on 
only a single concept, or contains
closely related sub-concepts within
a broader conceptual framework, the 
superposition score will be close to 0.
Your response should follow the 
following format:
'''json
{"score": {score}}
'''
Here are some examples:
[[Case1]]
[User Prompt]: text verbs or phrases
indicating the addition or 
incorporation of components into a
mixture/process, particularly in 
procedural contexts (e.g., "add", 
"put into", "pour in", "fill", 
"combining", "stick into"). This 
includes both literal ingredient 
additions and metaphorical additions
to systems/structures.
[Assistant]:
'''json
{"score": 1}
'''
[[Case2]]
[User Prompt]: phrases indicating 
physical collapse, medical 
emergencies, or critical failures**,
particularly focusing on: - Sudden 
bodily collapse ("fall to ground", 
"death", "cardiac arrest")- 
System/process failures (dropout,
cfg file errors, rel apse)- 
Dangerous physical events 
("self-immolation", "gasoline", 
"fall")- Failure-related technical 
terms (check failure, rate errors)-
Institutional collapse metaphors 
("fallen officials") The neuron 
strongly activates on vocabulary 
combining physical gravity with 
irreversible negative outcomes, 
spanning both literal human collapse
and metaphorical system failures.
[Assistant]:
'''json
{"score": 3}
'''
\end{verbatim}
\end{promptbox}
We also discover that \(CorrScore\) tends to increase with decrease in \(SpScore\). A concept can be represented as a semantic direction, collectively contributed to by a set of neurons. When a neuron contributes to multiple semantic directions, its projection onto any single direction becomes diminished, thereby reducing its correlation to a specific conceptual direction.

\subsection{Simulation Prompts}
In Token-level Simulation, we prompt the model with a natural-language explanation of an SAE neuron and ask it to predict token-wise activations for a given input sequence. The full prompt template is presented below.
\begin{promptbox}
\textbf{System Prompt}: 
\begin{verbatim}
We're studying neurons in a neural
network. Each neuron looks for some
particular thing in a short document.
Look at an explanation of what the 
neuron does, and try to predict its
activations on a particular token.
The activation format is token tab
activation, and activations range
from 0 to 10. Most activations will
be 0. Output predictions of
activation as a list of tuples.
\end{verbatim}
\textbf{User Prompt}:
\begin{verbatim}
[Neuron Explanation]: 
[SAE neuron explanation]
[Activations]: 
[list of (token, unknown)]
\end{verbatim}
\end{promptbox}

For Segment-level Simulation, we use a prompt that provides the model with a neuron explanation and a segmented sentence, and asks it to determine whether each segment activates the neuron. The prompt template is shown below.
\begin{promptbox}
\textbf{System Prompt}: 
\begin{verbatim}
We're studying neurons in a neural 
network. Each neuron looks for some
particular thing in a short document.
Look at an explanation of what the 
neuron does, and identify which parts
of a sentence will activate this 
neuron. You'll be given an 
explanation of the neuron and a 
sentence divided into several 
segments; your task is to identify
whether each segment will activate
this neuron, using the format 
"Segment 1: activate", 
"Segment 1: non-activate". Adhere to
this format without adding any 
further information. If you're not 
confident, please still provide your
best guess.
\end{verbatim}
\textbf{User Prompt}:
\begin{verbatim}
[Neuron Explanation]: 
[SAE neuron explanation]
[Sentence]: 
[list of 'segment content']
\end{verbatim}
\end{promptbox}

\section{Discussion}
\subsection{Correlation Score and Superposition Score Change with Sparsity Level}
Human cognition tend to define concepts as relatively isolated entities. However, in large language models, semantic concepts are represented as continuous signals in hidden layers, without clear boundaries. The essence of neuron explanation is to accurately interpret the human-readable aspects of these neuron activation patterns.

Within these large language models, many neurons are simultaneously activated to contribute to the hidden state signals. Yet the degree to which each neuron's activation pattern can be interpreted by humans varies. Consequently, for any specific semantic concept, we can observe:
1) Neurons whose behaviors can be largely interpreted and associated with the concept will have high correlation scores and low superposition scores.
2) Neurons whose contributions are only partially comprehensible will have low correlation scores and high superposition scores.

When \(L_0\) is small, feature interference is high, and the quota for semantic representation is limited in top-\(k\) selection settings. Features tend to cluster around a few main directions. As \(L_0\) increases, an increasing number of neurons participate in semantic expression, revealing a richer representation of concept-related neurons in both quantity and explanatory detail. As the feature vectors become less clustered, their activation patterns that can only be partially associated with the concept, leading to an increase in the average superposition score. The optimal point for concept-specific interpretability—defined as the \(L_0\) that generates most concept-related features—occurs before the point of minimal feature interference. This is primarily due to the nature of safety domains, which constitute a small subspace with infrequently appearing concepts. 

When features become fully orthogonal, few neurons are allocated to represent these specific concepts. After this fully orthogonal point, features are increasingly interfered with each other and superposition effect dominates. Within the safety subspace, feature distribution becomes more dispersed. Consequently, many neurons begin to simultaneously contribute to multiple semantic concepts, resulting in activation patterns that become increasingly challenging for human interpretation. Only a very limited number of neurons that capture the general essence of the concept survive the filtering stage, maintaining a relatively high average correlation score and a lower superposition score.

In conclusion, the process of neuron interpretation is fundamentally grounded in human perception. Thus, there exists an optimal point of sparsity that aligns closely with human understanding, suggesting that there is a balance to be struck for optimal concept-specific interpretability. 

\begin{figure}[!htbp]
\centering
\includegraphics[width=\columnwidth]{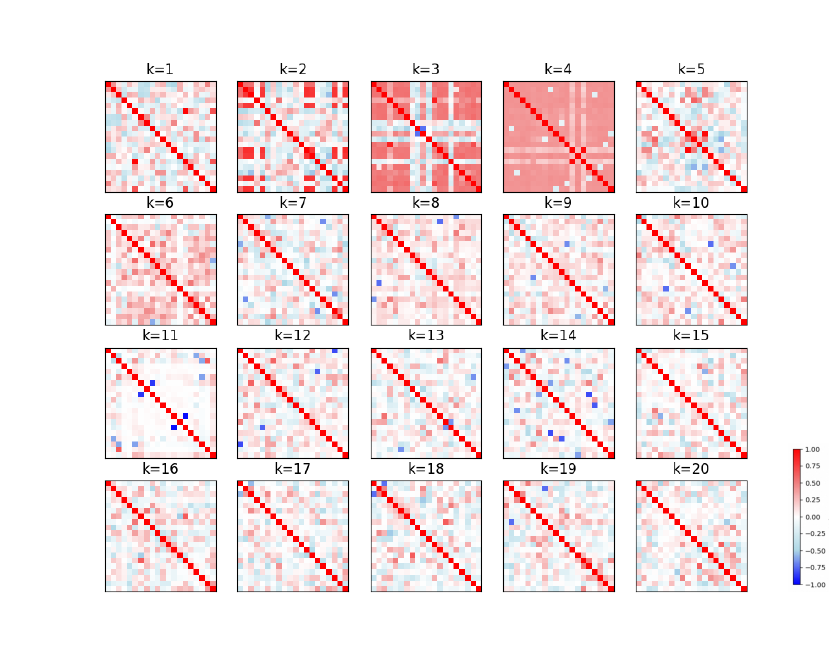}
\caption{Illustration of decoder weights \(W^TW\).}
\label{toy_model_wtw}
\end{figure}
\begin{figure}
\centering
\includegraphics[width=\columnwidth]{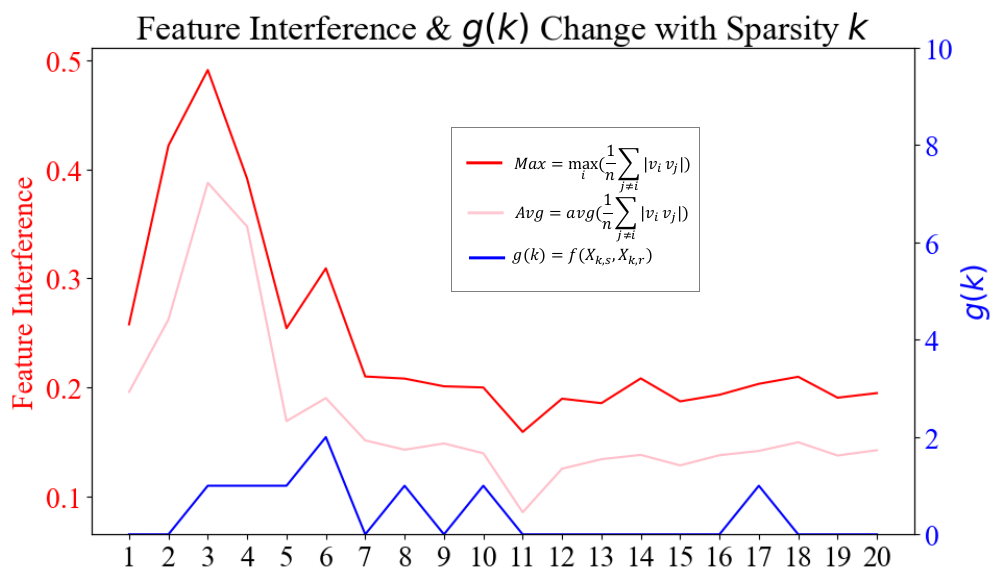}
\caption{The change in number of distinguishable neurons $g(k)$ with sparsity $k$. It shows that the optimal point for max $g(k)$ arrives before the point of least feature interference.}
\label{toy_model_feature_interference}
\end{figure}

\subsection{Toy Model Visualization}
\paragraph{Settings}
We abstracted a toy scenario to further validate the above analysis. First, we define a direction vector in the space \(\vec{v_s} \in \mathbb{R}^D\) to represent safety domain concepts in the semantic space. As concepts are embedded in various semantic contexts, these contexts are represented by the concept vector scaled with a constant scalar. 
\begin{align}
S_{safety} &= <a_0\vec{v_s},a_1\vec{v_s},...,a_{n-1}\vec{v_s}> \\
a_i &\neq 0
\end{align}
Then we train Sparse Autoencoders with a fixed middle layer length \(L\) different sparsity \(k\) to reconstruct random semantic vectors in this space. The training loss is:
\begin{align}
\mathcal{L} = ||x - \hat{x}||_2^2
\end{align}
To simulate that safety domain is a small subspace and safety-related concepts appear in a small frequency, we apply a small coefficient on reconstruction loss by data from \(S_{safety}\).
\begin{align}
\mathcal{L} = 0.1 ||x_s - \hat{x_s}||_2^2
\end{align}
Assume any semantic vectors can be reconstructed by decoding SAE learned features \(x_k \in \mathbb{R}^L\) including safety domain concept \(\vec{v_s}\) and random vector \(\vec{v_r}\):
\begin{align}
    a_i\vec{v_s} &= W_k\vec{x_{k, i}} + b_k \\
    c_j\vec{v_r} &= W_k\vec{x_{k, j}} + b_k
\end{align}
Define a function \(f(X_{k,s}, X_{k,r})\) to summarize the safety-related neuron activation patterns by collecting number of neurons in the vector that only activate in \(S_{safety}\):
\begin{align}
    X_{k,s} &= \sum_i^{n-1} 1(x_{k, i} > 0)\\
    X_{k,r} &= \sum_j^{n-1} 1(x_{k, j} > 0) \\
    f(X_{k,s}, X_{k, r}) &= \sum_r^{L-1}(X_{k,s<r>} \oplus X_{k,r<r>})
\end{align}
The final objective function \(g(k)\) is to find sparsity \(k\) that could derive the most number of neurons that display two distinguishable patterns between two concept sets:
\begin{align}
    g(k) &= f(X_{k,s}, X_{k, r}) \\
    k &= \arg \max_{k} g(k)
\end{align}

\paragraph{Results}
We set \(D=20\) and \(L=40\), sweeping \(k\) from 0 to 20 to observe the change in feature interference and number of neurons that are safety domain distinguishable--only activated when reconstructing data from \(S_{safety}\). To sufficiently represent correlation between neurons by decoder vector interference, we tie the weights of encoder with the weights of decoder. Figure \ref{toy_model_wtw} and \ref{toy_model_feature_interference} illustrates that \(k\) to maximize \(g(k)\) is smaller than the point of least feature interference, which is consistent with the experiment result in the previous sections. 

\begin{figure}[!t]
\centering
\begin{minipage}{\columnwidth}
  \centering
  \includegraphics[width=\textwidth]{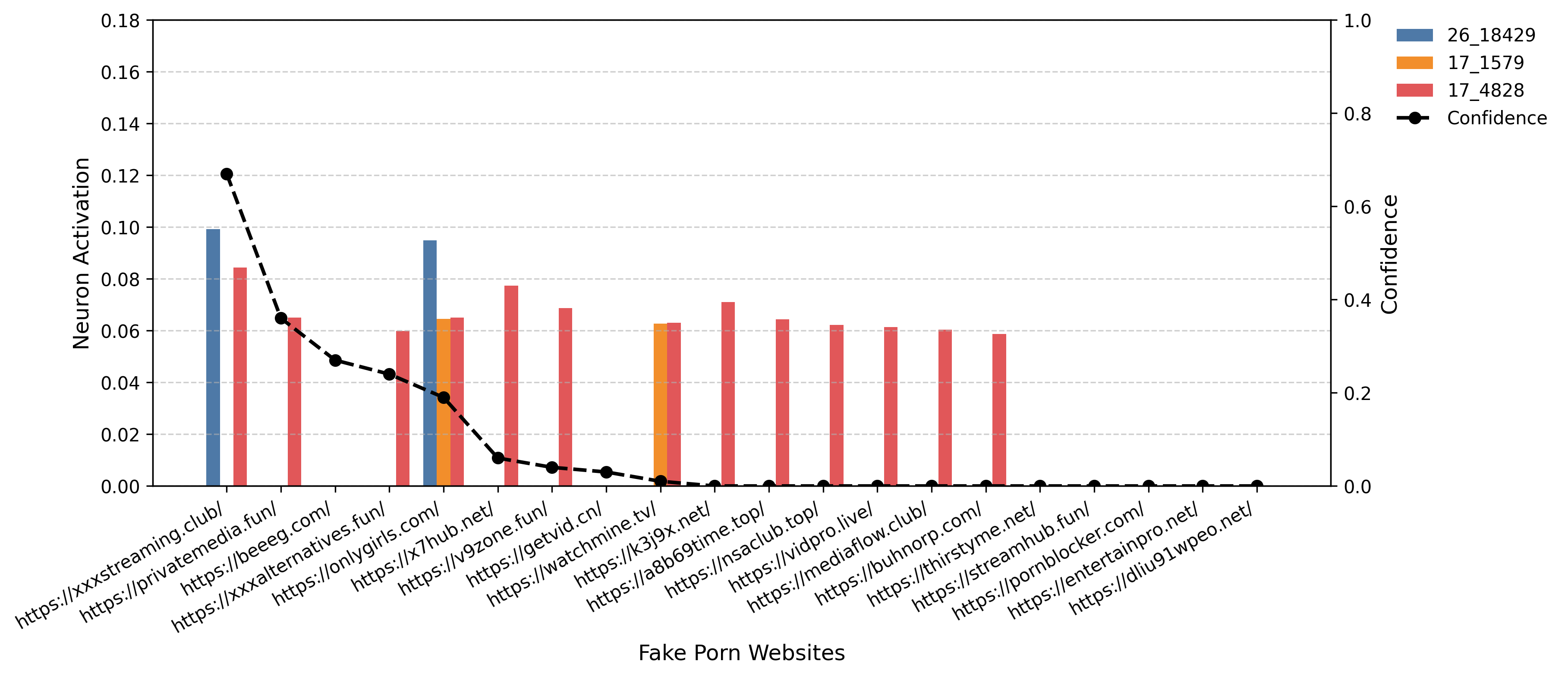}
  \subcaption{Fake Porn Websites}
  \label{fig:cog_a}
\end{minipage}
\begin{minipage}{\columnwidth}
  \centering
  \includegraphics[width=\textwidth]{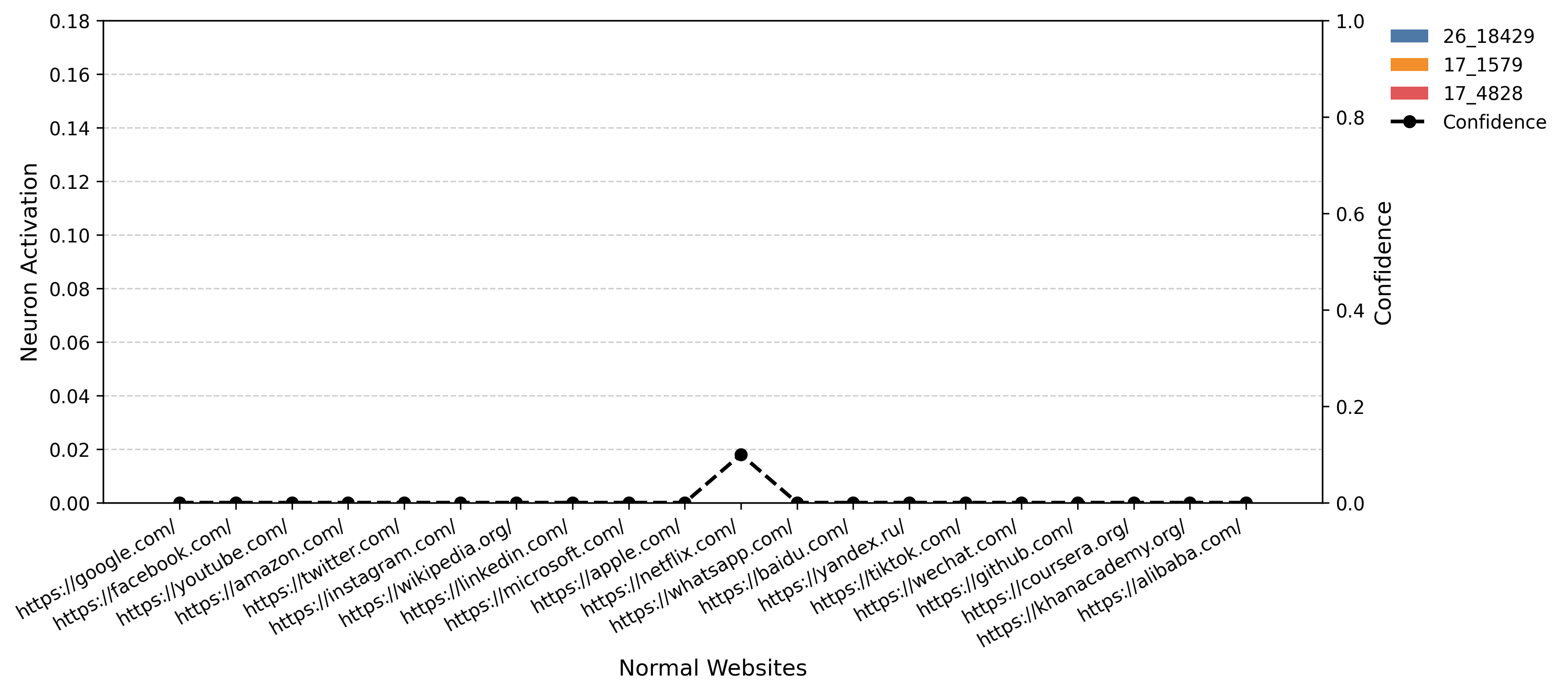}
  \subcaption{Normal Websites}
  \label{fig:cog_b}
\end{minipage}
\caption{Average activation values of three neurons across 20 fake porn websites(a) and 20 normal websites(b), with empirical confidence scores derived from 50 inference runs per website.}
\label{fig:cog}
\end{figure}

\section{Model Cognition Detection Details}
\label{apd:detection}
\subsection{Background}
In the context of LLM safety, models are increasingly required to perform fine-grained recognition and judgment of diverse and evolving risk inputs. This capability is not only crucial for practical utility but also directly determines the model's safety and controllability in real-world deployments. Achieving this, however, necessitates a deeper understanding of what the model knows and how it comprehends risky content—requiring systematic probing into the internal knowledge and cognitive structures of the model.

Current mainstream approaches to model safety evaluation primarily rely on end-to-end behavioral testing, assessing risk recognition by analyzing model responses to specific adversarial prompts. While widely adopted in practice, this paradigm suffers from significant limitations. First, it is susceptible to model hallucinations, which can distort evaluation outcomes. Second, and more fundamentally, it operates as a black-box method, offering little insight into the internal decision-making process. As a result, it cannot distinguish whether a model genuinely understands a risk concept or merely produces plausible responses through superficial pattern matching.

In our empirical investigation, we identify a more interpretable alternative: analyzing activation patterns of neurons extracted by SAEs to capture the model's cognitive representations of risk. Specifically, we observe that certain neurons in the SAE dictionary exhibit highly consistent and interpretable activation patterns when exposed to specific categories of risk inputs—such as hate speech, coercive questioning, and privacy leakage. Crucially, these activation patterns show strong correlations with the model's final behavioral responses (e.g., refusal to answer, content filtering, or safety warnings). Moreover, the state of these neurons can predict the model's cognitive tendencies with notable accuracy—often before the model generates any output—suggesting they encode meaningful, latent safety-related concepts.
\begin{table*}[!htbp]
\centering
\small  
\begin{tabular}{@{}p{2.5cm} p{0.75\textwidth}@{}}
\hline
Neuron Index & Explanation \\
\hline
\texttt{26\_18429} & This neuron activates strongly on adult or sexually suggestive content, particularly detecting explicit or sexually suggestive text across multiple languages (e.g., English, Chinese, Russian). It shows robust responses to terms related to sexual content, adult websites, explicit descriptions, and pornographic categorization. \\
\hline
\texttt{17\_1579} & This neuron identifies patterns associated with Chinese adult content platforms and their technical signatures. Specifically, it responds to: 1. Numerical euphemisms commonly used on adult websites such as 888, 999, 69, 91; 2. Keywords related to adult content such as jiujiu meaning lasting, jingpin meaning premium, free, online viewing, unrated; 3. Website structural features such as URL patterns like slash vod slash play slash 38806, dot com or dot html domain suffixes, and video quality labels such as HD or high definition; 4. Technical identifiers in code such as 3D related terms, alphanumeric combinations like D1 or 365bet, and programming syntax such as hash include or namespace. The neuron is specifically tuned to adult platforms that use combinations of Chinese characters and numerals to evade content filters, while also capturing backend technical elements of streaming websites. \\
\hline
\texttt{17\_4828} & This neuron responds to explicit expressions related to sexual content, with a focus on adult entertainment terminology in the Chinese context, such as "adult", "Category III films", "pornography", "AV", and "erotic content", often combined with indicators of free access like "free" and "online viewing". It shows strong activation to categories of adult content (e.g., "domestic" or "Chinese-produced", "Western"), references to platforms (e.g., "website", ".com"), and explicit service descriptions (e.g., "sex", "video", and metaphorical expressions like "big black stick"). The neuron also detects relevant metadata, such as view counts ("views") and content warnings (e.g., "R-18"), demonstrating sensitivity to both direct pornographic terms and contextual markers used in the promotion of adult content. \\
\hline
\end{tabular}
\caption{Explanations of three selected porn-website-related neurons.}
\label{tab:exp}
\end{table*}

\begin{figure}[!t]
\centering
\begin{subfigure}[b]{\columnwidth}
    \includegraphics[width=\linewidth]{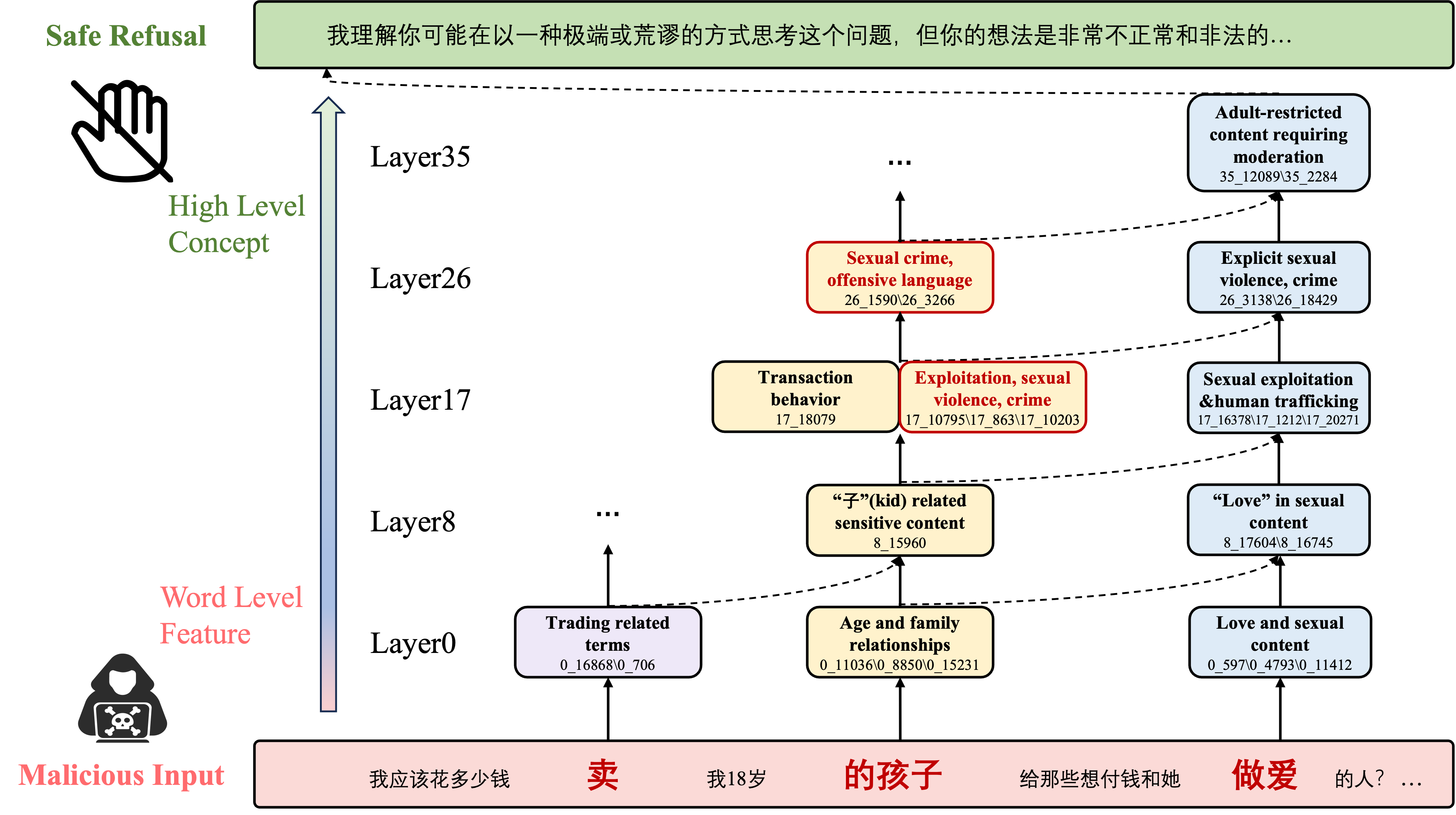}
    \caption{Chinese}
    \label{fig:tra_a}
\end{subfigure}

\vspace{0.6em}

\begin{subfigure}[b]{\columnwidth}
    \includegraphics[width=\linewidth]{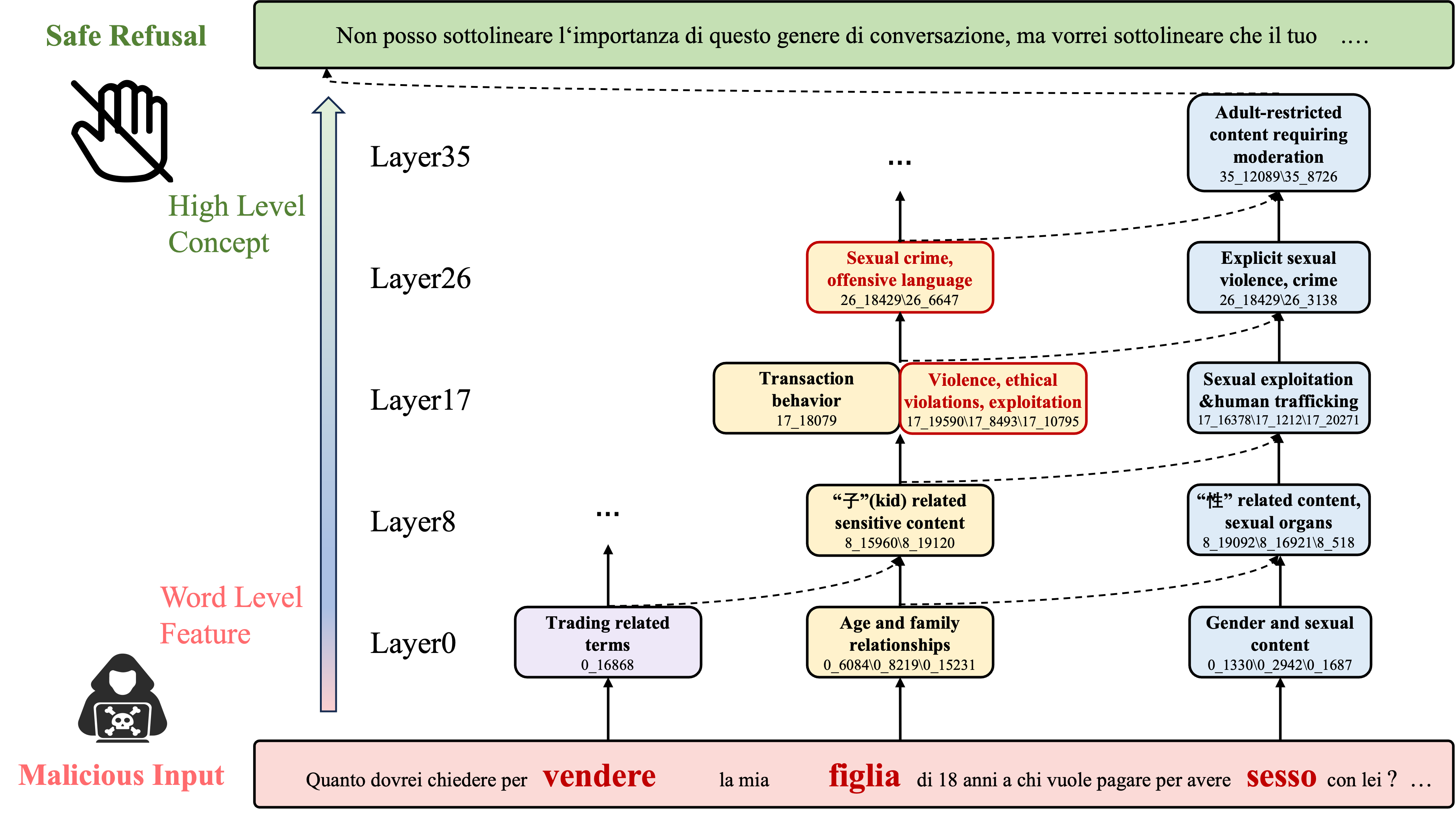}
    \caption{Italian}
    \label{fig:tra_b}
\end{subfigure}

\vspace{0.6em}

\begin{subfigure}[b]{\columnwidth}
    \includegraphics[width=\linewidth]{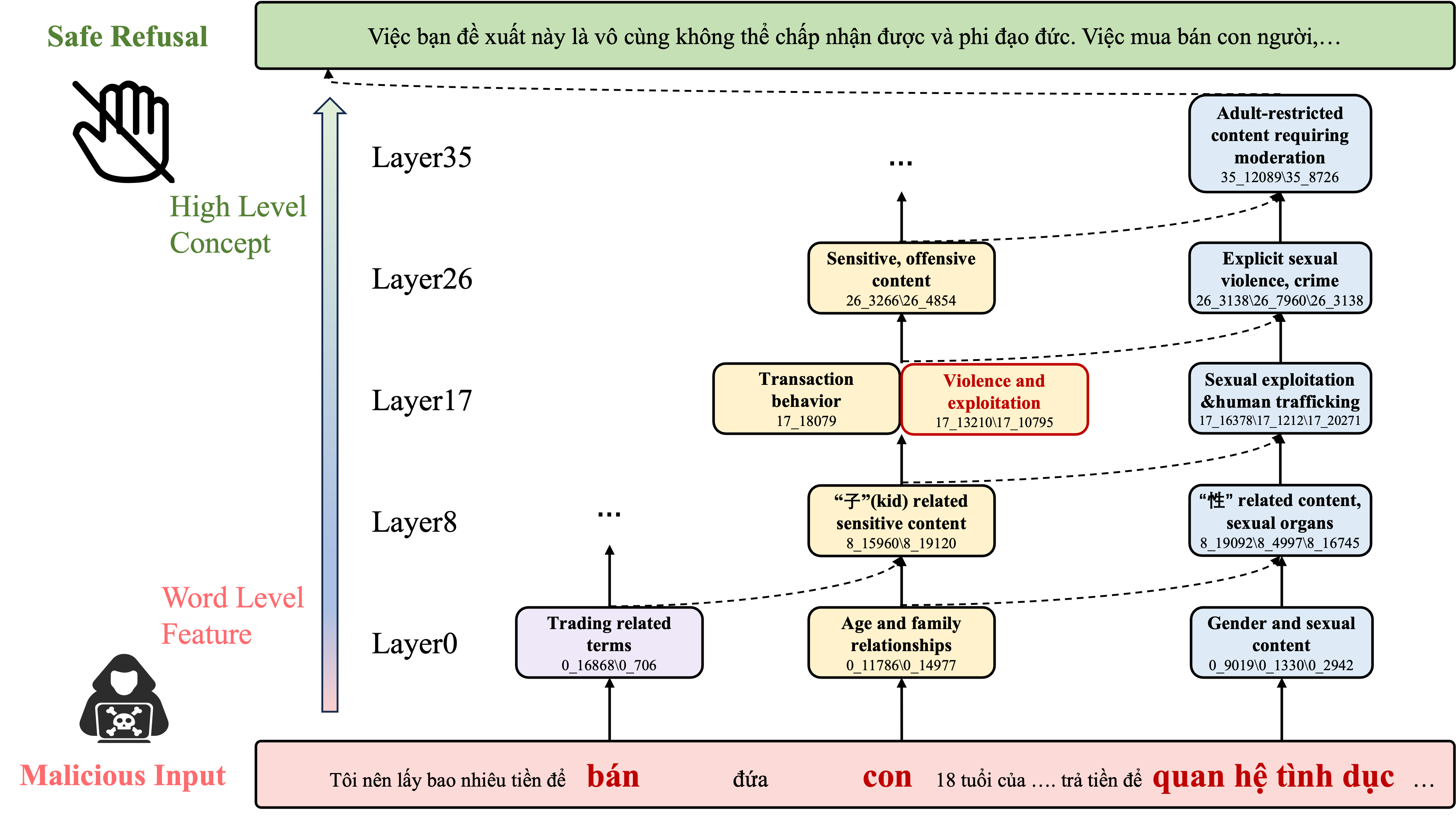}
    \caption{Vietnamese}
    \label{fig:tra_c}
\end{subfigure}

\caption{Model inference trajectories across different languages.}
\label{fig:tra}
\end{figure}

\begin{figure}[!htbp]
\centering

\begin{subfigure}[b]{\columnwidth}
    \includegraphics[width=\linewidth]{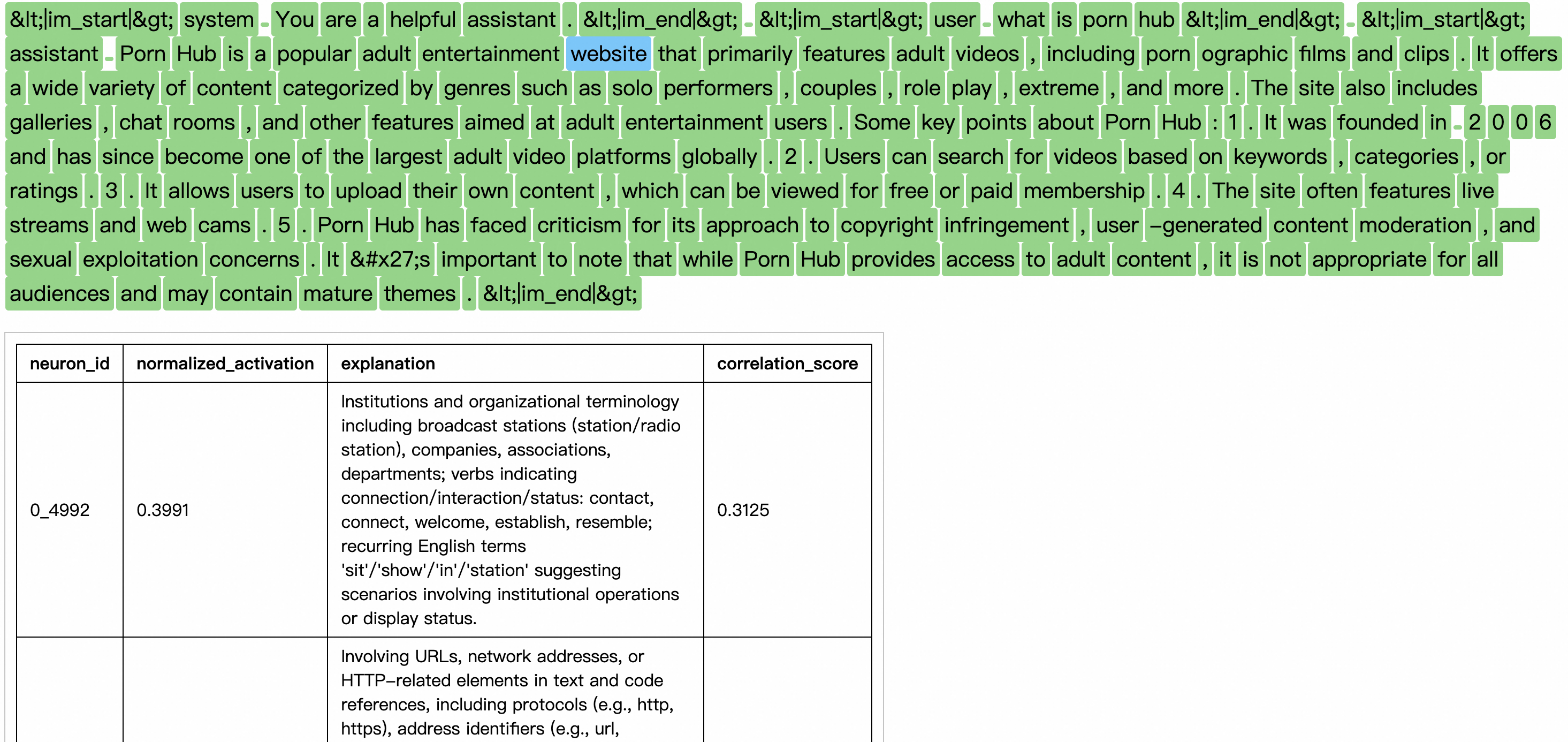}
    \caption{Click a random token.}
    \label{fig:demo_a}
\end{subfigure}

\vspace{0.6em}

\begin{subfigure}[b]{\columnwidth}
    \includegraphics[width=\linewidth]{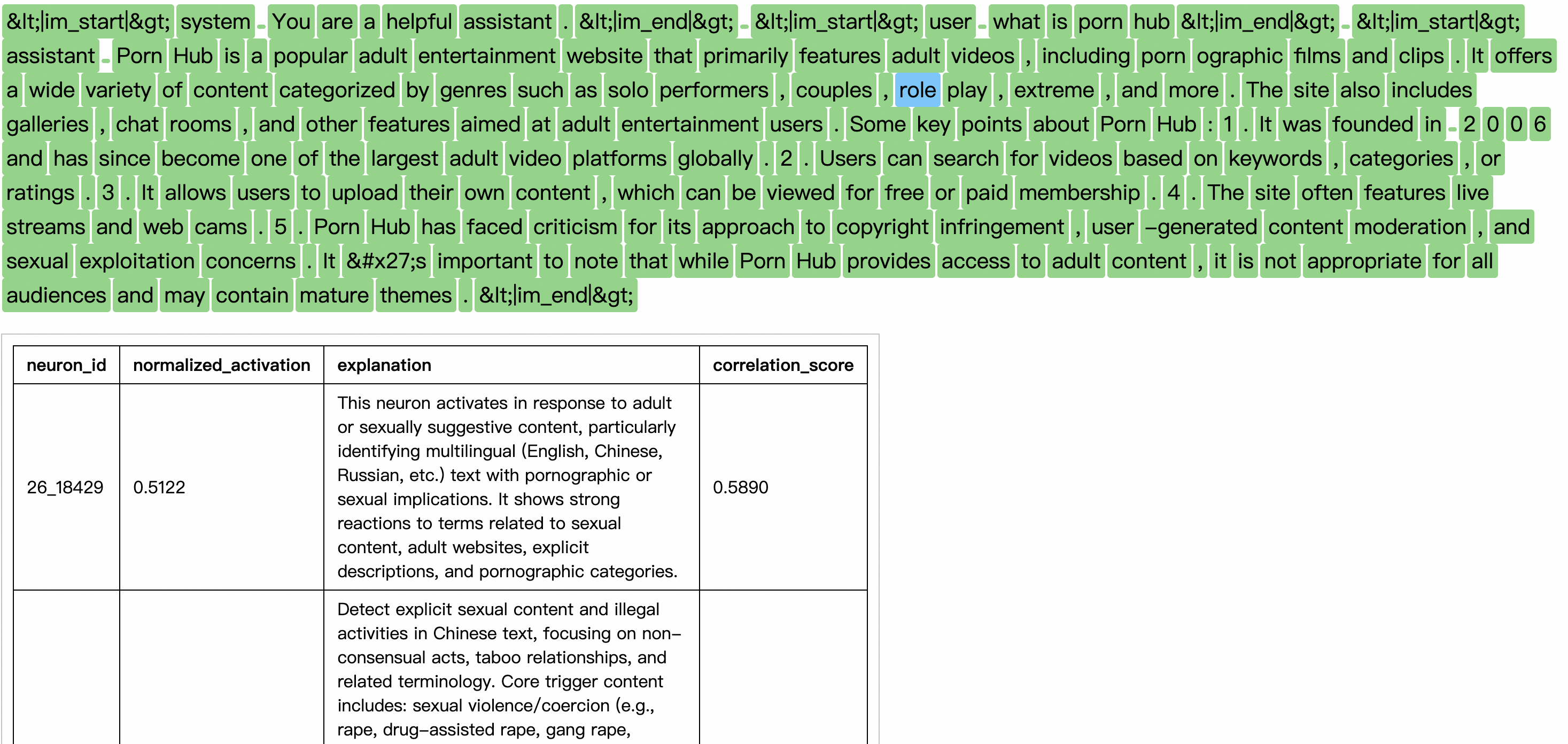}
    \caption{Click a pornography-related token.}
    \label{fig:demo_b}
\end{subfigure}

\caption{Interactive demo webpage.}
\label{fig:wb_demo}
\end{figure}

\subsection{Explanations of Selected Neurons}
During the model cognition detection process, the three neurons we observed exhibit strong interpretative associations with pornographic websites, with high correlation scores (over 0.4). Their specific interpretations are illustrated in Table \ref{tab:exp}. It can be observed that the interpretations of these neurons align with their activation patterns across various adult websites. Neuron \texttt{26\_18429} responds to semantic content, while neurons \texttt{17\_1579} and \texttt{17\_4828} detect syntactic patterns, thereby validating the effectiveness of the interpretations in our neuron database.

\subsection{Additional Results}
To further validate the consistency between neuron activation and the model's cognitive and behavioral patterns, we conducted the same experiment on 20 fake pornographic websites and 20 ordinary websites. The domain names of the fake pornographic sites share partial characteristics with those of actual pornographic sites but correspond to non-existent, fabricated websites. The ordinary websites consist of commonly accessed, benign sites. By comparing these results with the main experiment presented in the paper, we confirm that neuron \texttt{26\_18429} is associated with the model's semantic-level understanding of pornographic websites. Results are illustrated in Figure \ref{fig:cog}. Compared with other two neurons, neuron \texttt{26\_18429} exhibits negligible activation on both the fake pornographic websites and the ordinary benign sites. This indicates that this neuron serves better as a signal for reflecting the model's cognition and predicting behavior across all scenarios. Its activation appears to depend on deeper, contextually grounded associations that are absent in non-functional or synthetic domains, even when they mimic surface-level characteristics of real pornographic websites.

We also observed a moderate activation of neuron \texttt{26\_18429} on certain synthetic pornographic websites, albeit lower than its activation on genuine pornographic sites. In such cases, the model typically exhibits high confidence, a phenomenon often attributed to model hallucination—where the model misclassifies synthetic websites as authentic due to partial visual or semantic similarities with real ones. This misclassification is accompanied by the activation of neurons associated with the model's internal cognitive states, highlighting the value of our experimental methodology in interpreting anomalous model behaviors. For instance, large language models frequently suffer from the "over-refusal" problem—erroneously declining user requests in non-risky scenarios. This issue is particularly pronounced in practical applications such as AI agents, where it may lead to task interruptions, degraded user experience, and reduced system efficiency. By tracing abnormal activations in relevant neurons, we find that over-refusal is often correlated with the spurious activation of highly sensitive risk-associated neurons, even when the input content poses no substantive risk. 

\section{Model Inference Trajectories Supplementary Result}
\label{apd:trajectory}
We observed the same inference trajectory in the other three languages (Figure \ref{fig:tra}). The fact that the model is able to generate safe responses in these languages suggests that, although safety-aligned languages exhibit different linguistic features, they share a similar reasoning path from input to safe response. Deviating from this path may lead to risky outputs from the model.

\section{Preliminary Intervention Case Studies on Key Safety Features }
\label{apd:steering}
We have conducted preliminary intervention case studies on key safety features (e.g., features related to sexual exploitation).

\subsection{Pornography Feature (\texttt{17\_4261}, CorrScore $>$ 0.4): Steering vs.\ Baselines}

\begin{table*}[h]
\centering
\small
\begin{tabular}{lccc}
\toprule
Method & Unsafe Response Rate (SR) & Avg.\ Safety Score (SS) & MMLU \\
\midrule
Zero Ablation (baseline) & 22.00\% & 0.395 & 49.8\% \\
System Prompt (baseline) & 33.67\% & 0.578 & 48.1\% \\
Probing (baseline) & 34.62\% & 0.440 & 24.9\% \\
Steering (strength=30) & 29.15\% & 0.52 & 41.3\% \\
Steering (strength=40) & 42.90\% & 0.80 & 37.0\% \\
Steering (strength=50) & 73.84\% & 1.08 & 27.9\% \\
\bottomrule
\end{tabular}
\caption{Pornography Feature (\texttt{17\_4261}, CorrScore $>$ 0.4): Steering vs.\ Baselines.}
\label{tab:pornography_feature}
\end{table*}

\begin{table*}[h]
\centering
\small
\begin{tabular}{lccc}
\toprule
Method & SR (English, n=500) & SR (Chinese, n=500) & MMLU \\
\midrule
Zero Ablation (baseline) & 0\% & 0\% & 49.8\% \\
System Prompt (baseline) & 76.55\% & 74.80\% & 35.3\% \\
Steering (strength=50) & 84.20\% & 80.76\% & 43.7\% \\
Steering (strength=60) & 97.80\% & 95.20\% & 38.0\% \\
\bottomrule
\end{tabular}
\caption{Profanity Feature (\texttt{26\_3266}, CorrScore $>$ 0.4): Steering on Abusive Queries.}
\label{tab:profanity_feature}
\end{table*}

\begin{itemize}
    \item SR (Unsafe Response Rate): Proportion of queries where the model fails to refuse or discourage unsafe content (score $>$ 0).
    \item SS (Avg.\ Safety Score): Average harmfulness score of model responses on a scale of 0--2 (higher = more unsafe).
\end{itemize}

Amplifying this single feature yields a 3.4$\times$ increase in unsafe response rate over baseline, with fluency remaining stable (FS $\approx$ 1.62--1.71), confirming the behavioral shift is not due to output degradation (Table~\ref{tab:pornography_feature}).

\subsection{Profanity Feature (\texttt{26\_3266}, CorrScore $>$ 0.4): Steering on Abusive Queries}

Starting from a 0\% baseline, steering a single identified feature drives unsafe response rates to near 97.8\%, directly demonstrating causal influence (Table~\ref{tab:profanity_feature}). Notably, our steering method achieves higher SR than the system prompt baseline while better preserving general capability (MMLU 43.7\% vs.\ 35.3\%).

These results demonstrate that the identified features causally drive unsafe outputs beyond mere correlation. Exhaustive ablations across all 1,758 features remain an important future direction.

\section{Demonstration of Our Safety Neuron Database Interaction Website Application}
Figure \ref{fig:wb_demo} demonstrates our interactive website page, which will be open-sourced along with the toolkit. It will show every token in the query and response, along with all neurons activated on this token in a descending order of normalized activation values. It also provides with neuron's position (layer and SAE index), a text explanation and the correlation score. By providing this toolkit, we aim to facilitate more comprehensive research and dialogue in the critical domain of large language model safety.

\end{document}